\documentclass[runningheads]{llncs}
\usepackage{graphicx}
\usepackage{amsmath,amssymb} %
\usepackage{color}
\usepackage[width=122mm,left=12mm,paperwidth=146mm,height=193mm,top=12mm,paperheight=217mm]{geometry}

\usepackage{etoolbox}
\patchcmd{\paragraph}{\itshape}{\bfseries\boldmath}{}{}

\usepackage[sort]{cite}

\usepackage[colorlinks=true]{hyperref}
\usepackage{subcaption}
\usepackage{alex}
\usepackage{tikz,pgfplots}
\usepackage{tango}
\usepackage{nicefrac}
\usepackage{mathtools}
\DeclarePairedDelimiter{\ceil}{\lceil}{\rceil}

\newcommand{\reffig}[1]{Figure~\ref{fig:#1}}

\newcommand{\refsec}[1]{Section~\ref{sec:#1}}

\newcommand{\lblfig}[1]{\label{fig:#1}}
\newcommand{\lblsec}[1]{\label{sec:#1}}

\newcommand{\mv}{\mathcal{T}}
\newcommand{\res}{\mathcal{R}}

\pgfplotsset{compat = 1.3,
          legend style={font=\scriptsize},
          legend cell align={left},
          legend style={cells={align=left}, draw=black!20},
          grid=both,grid style={dotted},tick style={draw=none},
          enlarge x limits=false,
          enlarge y limits=false,
          axis line style={draw=black!20},}
\pgfplotscreateplotcyclelist{mystylelist}{%
densely dashdotted\\
solid\\
densely dashed\\
loosely dotted\\
dashed\\
dashdotted\\
densely dotted\\
}
\pgfplotscreateplotcyclelist{mycolorlist}{%
chameleon3\\
skyblue2\\
scarletred3\\
aluminium6\\
plum1\\
orange1\\
skyblue3\\
chocolate2\\
maroon\\
butter3\\
}
\pgfplotscreateplotcyclelist{mycolorlist2}{%
butter1\\
chameleon3\\
skyblue2\\
plum1\\
aluminium6\\
orange1\\
scarletred3\\
skyblue3\\
chocolate2\\
maroon\\
}
\pgfplotscreateplotcyclelist{mystylelist2}{%
dash dot dot\\
densely dashdotted\\
solid\\
loosely dotted\\
densely dashed\\
dashed\\
dashdotted\\
densely dotted\\
}
\pgfplotscreateplotcyclelist{mycolorlist3}{%
butter1\\
chameleon3\\
skyblue2\\
plum1\\
aluminium6\\
scarletred3\\
skyblue3\\
chocolate2\\
maroon\\
}
\pgfplotscreateplotcyclelist{mystylelist3}{%
dash dot dot\\
densely dashdotted\\
solid\\
loosely dotted\\
densely dashed\\
dashdotted\\
densely dotted\\
}
\pgfplotscreateplotcyclelist{mycolorlistdark}{%
scarletred3\\
aluminium6\\
plum1\\
skyblue2\\
chameleon3\\
orange1\\
skyblue3\\
chocolate2\\
maroon\\
butter3\\
}

\pgfplotsset{
    layers/my layer set/.define layer set={
      background,
      main,
      foreground
    }{
    },
    set layers=my layer set,
}

\begin{document}
\pagestyle{headings}
\mainmatter

\title{Video Compression through Image Interpolation} %

\titlerunning{Video Compression through Image Interpolation}

\authorrunning{Chao-Yuan Wu, Nayan Singhal, Philipp Kr\"ahenb\"uhl}

\author{Chao-Yuan Wu, Nayan Singhal, Philipp Kr\"ahenb\"uhl}
\institute{The University of Texas at Austin\\
  \email{ \{cywu, nayans, philkr\}@cs.utexas.edu}}

\maketitle

\begin{abstract}
An ever increasing amount of our digital communication, media consumption, and content creation revolves around videos.
We share, watch, and archive many aspects of our lives through them,
all of which are powered by strong video compression.
Traditional video compression is laboriously hand designed and hand optimized.
This paper presents an alternative in an end-to-end deep learning codec.
Our codec builds on one simple idea: Video compression is repeated image interpolation.
It thus benefits from recent advances in deep image interpolation and generation.
Our deep video codec outperforms today's prevailing codecs, such as H.261, MPEG-4 Part 2, and performs on par with H.264.
\end{abstract}

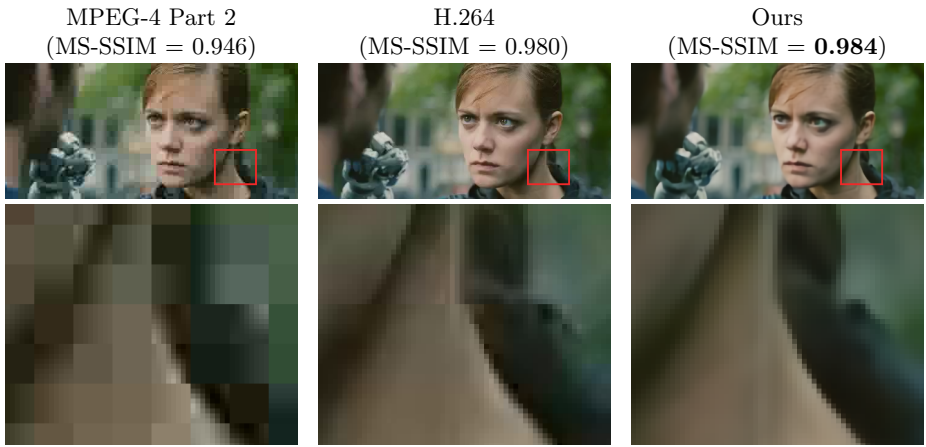
\begin{figure}[t]
\setlength{\tabcolsep}{0.1em}
\center
  \definecolor{tango_red}{RGB}{239,41,41}
  \begin{tabular}{ccc}
        MPEG-4 Part 2&H.264&Ours\\
        (MS-SSIM = 0.946)& (MS-SSIM = 0.980)&(MS-SSIM = {\bf 0.984})\vspace{-0.1cm}\\
    \begin{tikzpicture}\node { \includegraphics[width=0.32\linewidth,trim={1cm 0 1cm 0},clip]{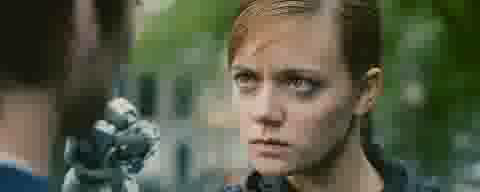} };\draw[draw=tango_red,line width=0.25mm] (0.85,-0.25) rectangle (1.4,-0.7);\end{tikzpicture}&
    \begin{tikzpicture}\node { \includegraphics[width=0.32\linewidth,trim={1cm 0 1cm 0},clip]{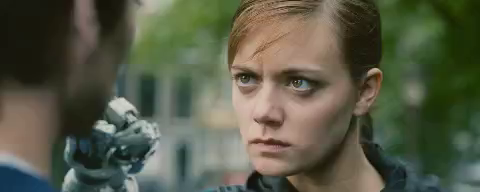} };\draw[draw=tango_red,line width=0.25mm] (0.85,-0.25) rectangle (1.4,-0.7);\end{tikzpicture}&
    \begin{tikzpicture}\node { \includegraphics[width=0.32\linewidth,trim={1cm 0 1cm 0},clip]{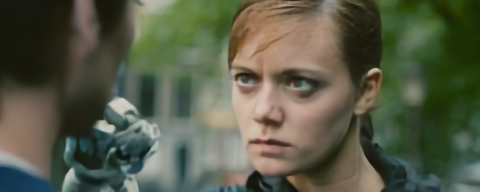} };\draw[draw=tango_red,line width=0.25mm] (0.85,-0.25) rectangle (1.4,-0.7);\end{tikzpicture}\vspace{-0.15cm}\\
        \includegraphics[width=0.32\linewidth,trim={0 20cm 0 20cm},clip]{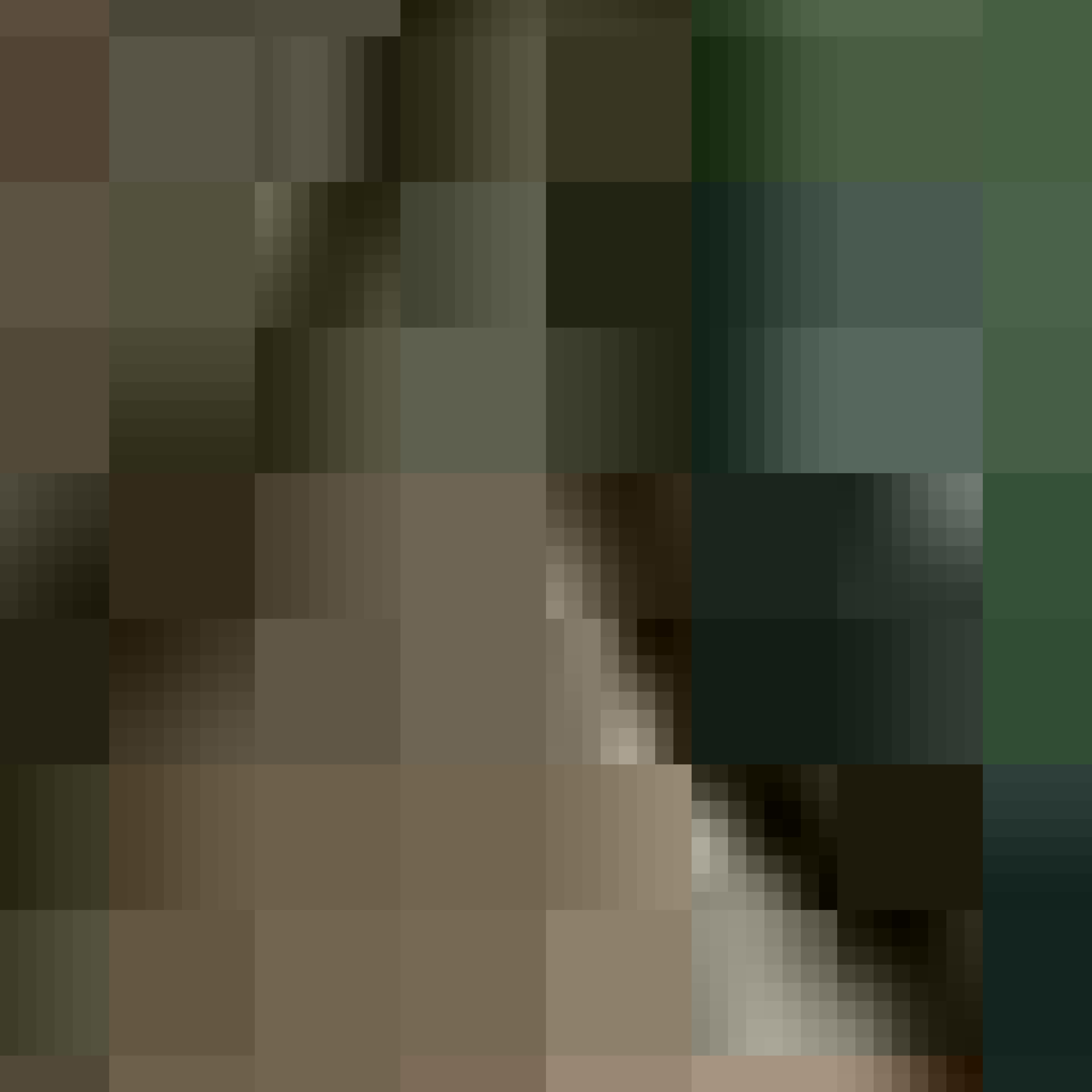}&\includegraphics[width=0.32\linewidth,trim={0 20cm 0 20cm},clip]{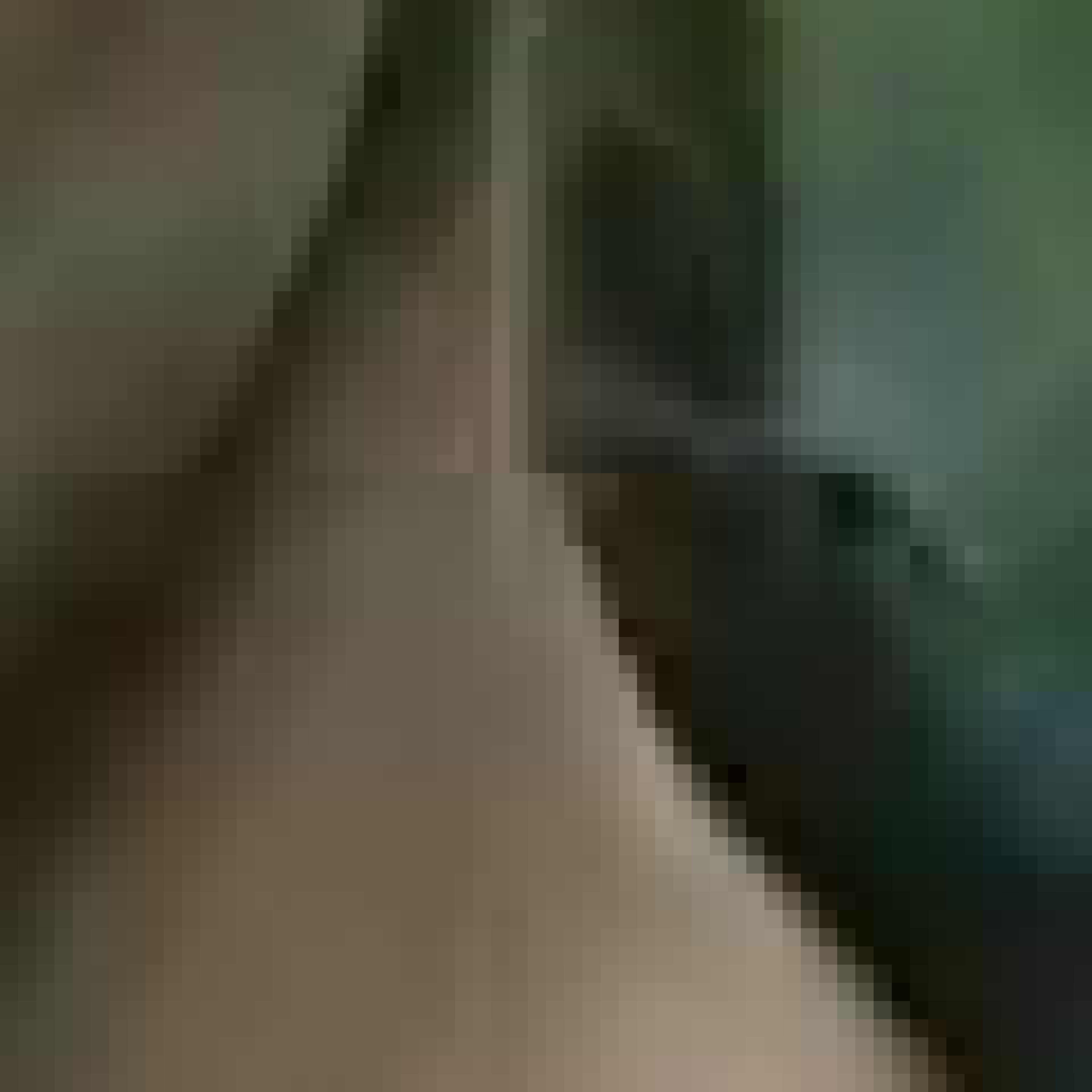}&\includegraphics[width=0.32\linewidth,trim={0 20cm 0 20cm},clip]{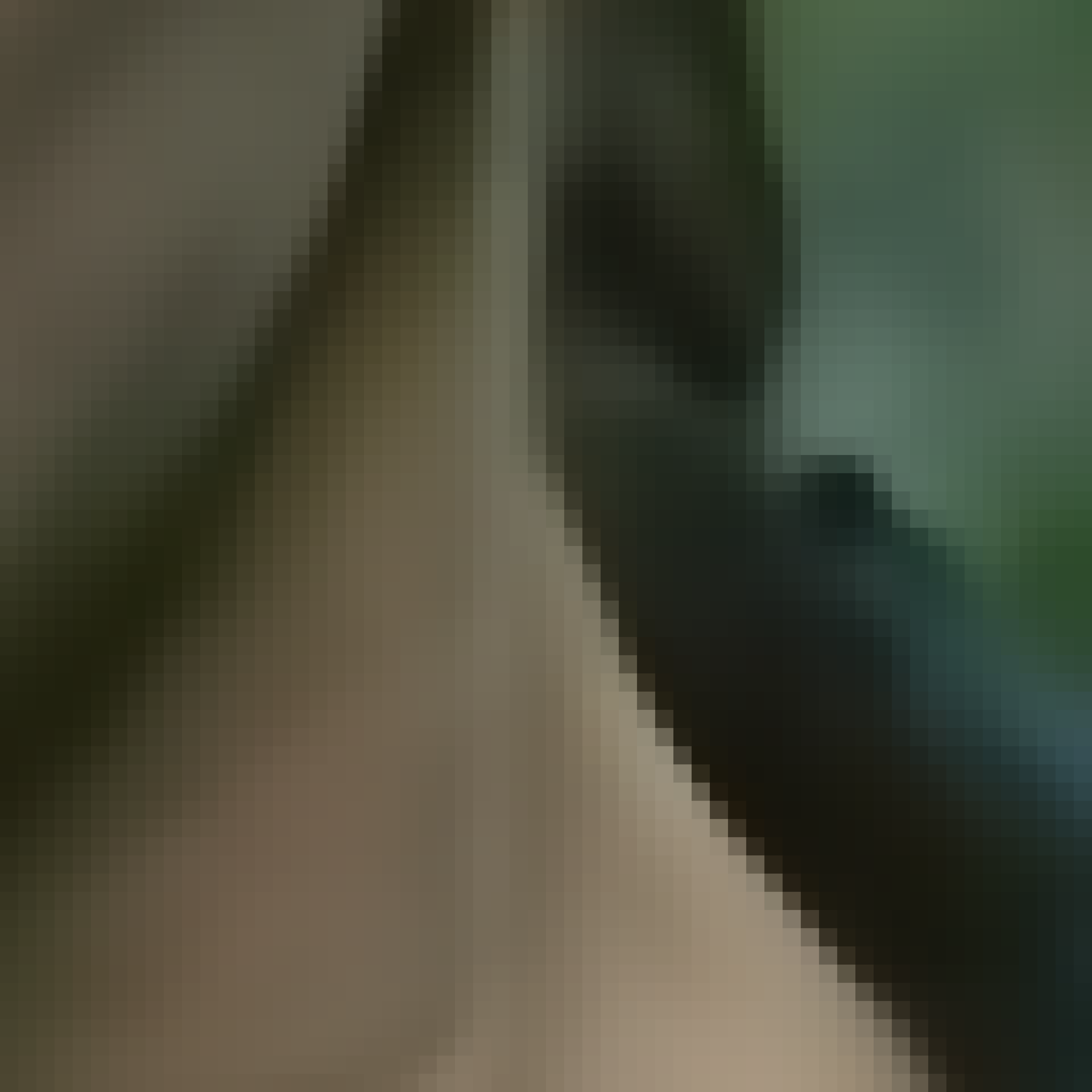}\vspace{-0.2cm}
  \end{tabular}
  \vspace{-2mm}
  \caption{Comparison of our end-to-end deep video compression algorithm to MPEG-4 Part 2 and H.264 on the Blender \textit{Tears of Steel} movie.
       All methods use 0.080 BPP.
       Our model offers a visual quality better than MPEG-4 Part 2 and comparable to H.264.
      Unlike traditional methods, our method is free of block artifacts.
      The MS-SSIM \cite{wang2003multiscale} measures the image quality of the video clip compared to the raw uncompressed ground truth.
      (Best viewed on screen.)}\vspace{-0.5cm}
  \label{fig:teaser}
\end{figure}

\section{Introduction}

Video commands the lion's share of internet data, and today makes up three-fourths of all internet traffic~\cite{cisco}.
We capture moments, share memories, and entertain one another through moving pictures, all of which are powered by ever powerful digital camera and video compression.
Strong compression significantly reduces internet traffic, saves storage space, and increases throughput.
It drives applications like cloud gaming, real-time high-quality video streaming~\cite{richardson2002video}, or 3D and 360-videos.
Video compression even helps better understand and parse videos using deep neural networks~\cite{coviar}.
Despite these obvious benefits, video compression algorithms are still largely hand designed.
The most competitive video codecs today rely on a sophisticated interplay between block motion estimation, residual color patterns, and their encoding using discrete cosine transform and entropy coding~\cite{schwarz2007overview}.
While each part is carefully designed to compress the video as much as possible, the overall system is not jointly optimized, and has largely been untouched by end-to-end deep learning.

This paper presents, to the best of our knowledge, the first end-to-end trained deep video codec.
The main insight of our codec is a different view on video compression:
We frame video compression as repeated image interpolation, and draw on recent advances in deep image generation and interpolation.
We first encode a series of anchor frames (key frames), using standard deep image compression.
Our codec then reconstructs all remaining frames by interpolating between neighboring anchor frames.
However, this image interpolation is not unique.
We additionally provide a small and compressible code to the interpolation network to disambiguate different interpolations, and encode the original video frame as faithfully as possible.
The main technical challenge is the design of a compressible image interpolation network.

We present a series of increasingly powerful and compressible encoder-decoder architectures for image interpolation.
We start by using a vanilla U-net interpolation architecture~\cite{ronneberger2015u}
for reconstructing frames other than the key frames.
This architecture makes good use of repeating static patterns through time,
but it struggles to properly disambiguate the trajectories for moving patterns.
We then directly incorporate an offline motion estimate from either block-motion estimation or optical flow into the network.
The new architecture interpolates spatial U-net features using the pre-computed motion estimate, 
and improves compression rates by an order of magnitude over deep image compression.
This model captures most, but not all of the information we need to reconstruct a frame.
We additionally train an encoder that extracts the content not present in either of the source images, and represents it compactly.
Finally, we reduce any remaining spatial redundancy, and compress them using a 3D PixelCNN~\cite{oord2016pixel} with adaptive arithmetic coding~\cite{witten1987arithmetic}.

To further reduce bitrate, our video codec applies image interpolation in a hierarchical manner.
Each consecutive level in the hierarchy interpolates between ever closer reference frames, and is hence more compressible.
Each level in the hierarchy uses all previously decompressed images.

We compare our video compression algorithm to state-of-the-art video compression (HEVC, H.264, MPEG-4 Part 2, H.261), and various image interpolation baselines.
We evaluate all algorithms on two standard datasets of uncompressed video: Video Trace Library (VTL)~\cite{vtl} and Ultra Video Group (UVG)~\cite{uvg}.
We additionally collect a subset of the Kinetics dataset~\cite{carreira2017quo} for both training and testing.
The Kinetics subset contains high resolution videos, which we down-sample to remove compression artifacts introduced by prior codecs on YouTube.
The final dataset contains 2.8M frames.
Our deep video codec outperforms all deep learning baselines, MPEG-4 Part 2, and H.261 in both compression rate and visual quality measured by MS-SSIM~\cite{wang2003multiscale} and PSNR.
We are on par with the state-of-the-art H.264 codec.
\figref{teaser} shows a visual comparison. 
All the data is publicly available, and we will publish our code upon acceptance.

\section{Related Work}
Video compression algorithms must specify an encoder for compressing the video, and a decoder for reconstructing the original video.
The encoder and the decoder together constitute a codec.
A codec has one primary goal: Encode a series of images in the fewest number of bits possible.
Most compression algorithms find a delicate trade-off between compression rate and reconstruction error.
The simplest codecs, such as motion JPEG or GIF, encode each frame independently, and heavily rely on image compression.

\vspace{-2mm}
\paragraph{Image compression.}

For images, deep networks yield state-of-the-art compression ratios with impressive reconstruction quality~\cite{waveone2017,johnston2017improved,toderici2017full,balle2016end,Theis2017a}.
Most of them train an autoencoder with a small binary bottleneck layer to directly minimize distortion~\cite{waveone2017,toderici2017full,johnston2017improved}.
A popular variant progressively encodes the image using a recurrent neural network~\cite{toderici2017full,baig2017learning,johnston2017improved}.
This allows for variable compression rates with a single model.
We extend this idea to variable rate video compression.

Deep image compression algorithms use fully convolutional networks to handle arbitrary image sizes.
However, the bottleneck in fully convolutional networks still contains spatially redundant activations.
Entropy coding further compresses this redundant information~\cite{mentzer2018conditional,waveone2017,toderici2017full,balle2016end,Theis2017a}.
We follow Mentzer~\etal~\cite{mentzer2018conditional} and use adaptive arithmetic coding on probability estimates of a Pixel-CNN~\cite{oord2016pixel}.

Learning the binary representation is inherently non-differentiable, which complicates gradient based learning.
Toderici \etal~\cite{toderici2017full} use stochastic binarization and backpropagate the derivative of the expectation.
Agustsson \etal~\cite{agustsson2017soft} use soft assignment to approximate quantization.
Balle \etal~\cite{balle2016end} replace the quantization by adding uniform noise.
All of these methods work similarly and allow for gradients to flow through the discretization.
In this paper, we use stochastic binarization~\cite{toderici2017full}.

Combining this bag of techniques, deep image compression algorithms offer a better compression rate than hand-designed algorithms, such as JPEG or WebP~\cite{webp}, at the same level of image quality~\cite{waveone2017}.
Deep image compression algorithms heavily exploit the spatial structure of an image.
However, they miss out on a crucial signal in videos: time.
Videos are temporally highly redundant.
No deep image compression can compete with state-of-the-art (shallow) video compression, which exploits this redundancy.

\vspace{-2mm}
\paragraph{Video compression.}
Hand-designed video compression algorithms, such as H.263, H.264 or HEVC (H.265)~\cite{le1991mpeg} build on two simple ideas: They decompose each frame into blocks of pixels, known as macroblocks, and they divide frames into image (I) frames and referencing (P or B) frames.
I-frames directly compress video frames using image compression.
Most of the savings in video codecs come from referencing frames.
P-frames borrow color values from preceding frames.
They store a motion estimate and a highly compressible difference image for each macroblock.
B-frames additionally allow bidirectional referencing, as long as there are no circular references.
Both H.264 and HEVC encode a video in a hierarchical way.
I-frames form the top of the hierarchy.
In each consecutive level, P- or B-frames reference decoded frames at higher levels.
The main disadvantages of traditional video compression is the intensive engineering efforts required
and the difficulties in joint optimization.
In this work, we build a hierarchical video codec using deep neural networks.
We train it end-to-end without any hand-engineered heuristics or filters.
Our key insight is that referencing (P or B) frames are a special case of image interpolation.

Learning-based video compression is largely unexplored, in part due to difficulties in modeling temporal redundancy. 
Tsai \etal propose a deep post-processing filter encoding errors of H.264 in domain specific videos~\cite{tsai2017learning}.
However, it is unclear if and how the filter generalizes in an open domain.
To the best of our knowledge, this paper proposes the first general deep network for video compression.

\vspace{-2mm}
\paragraph{Image interpolation and extrapolation.}
Image interpolation seeks to hallucinate an unseen frame between two reference frames.
Most image interpolation networks build on an encoder-decoder network architecture to move pixels through time~\cite{jia2016dynamic,niklaus2017sepvideo,lyty-vfsdv-17,jiang2017super}.
Jia~\etal~\cite{jia2016dynamic} and Niklaus~\etal~\cite{niklaus2017sepvideo} estimate a spatially-varying convolution kernel.
Liu~\etal~\cite{lyty-vfsdv-17} produce a flow field.
All three methods then combine two predictions, forward and backward in time, to form the final output.

Image extrapolation is more ambitious and predicts a future video from a few frames \cite{mcl-dmvpb-16}, or a still image \cite{vpt-gvsd-16, xue2016visual}. 
Both image interpolation and extrapolation works well for small timesteps, e.g. for creating slow-motion video \cite{jiang2017super} or predicting a fraction of a second into the future. %
However, current methods struggle for larger timesteps, where the interpolation or extrapolation is no longer unique, and additional side information is required.
In this work, we extend image interpolation and incorporate few compressible bits of side information to reconstruct the original video.

\section{Preliminary}

Let $I^{(t)} \in \RR^{W\times H\times 3}$ denote a series of frames for $t \in \{0,1,\ldots\}$.
Our goal is to compress each frame $I^{(t)}$ into a binary code $b^{(t)} \in \cbr{0, 1}^{N_t}$.
An encoder $E: \{I^{(0)},I^{(1)},\ldots\} \to \{b^{(0)},b^{(1)},\ldots\}$ and decoder $D:\{b^{(0)},b^{(1)},\ldots\} \to \{\hat I^{(0)},\hat I^{(1)},\ldots\}$ compress and decompress the video respectively.
$E$ and $D$ have two competing aims:
Minimize the total bitrate ${\sum_t N_t}$, and reconstruct the original video as faithfully as possible.
We measure the reconstruction error with an L1 loss $\ell(\hat{I}, I) = \|\hat{I} - I\|_1$.

\vspace{-2mm}
\paragraph{Image compression.}
The simplest encoders and decoders process each image independently: $E_I: I^{(t)} \to b^{(t)}$, $D_I:b^{(t)} \to \hat I^{(t)}$.
Here, we build on the model of Toderici~\etal~\cite{toderici2017full}, which encodes and reconstructs an image progressively over $K$ iterations. 
At each iteration, the model encodes a residual $r_k$ between the previously coded image and the original frame: 
\begin{align}
r_{0} &:= I\notag\\
b_{k} &:= E_I\rbr{r_{k-1}, g_{k-1}}, & & & r_{k} &:= r_{k-1} - D_I\rbr{b_k, h_{k-1}}, &&& \text{for } k = 1, 2, \dots\notag
\end{align}
where $g_k$ and $h_k$ are latent Conv-LSTM states that are updated at each iteration.
All iterations share the same network architecture and parameters forming a recurrent structure.
The training objective minimizes the distortion at all the steps ${\sum_{k=1}^K \|r_{k}\|_1}$.
The reconstruction $\hat I_K = \sum_{k=1}^K D_I(b_k)$ allows for a variable bitrate encoding depending on the choice of $K$.

Both the encoder and the decoder consist of $4$ Conv-LSTMs with stride $2$.
The bottleneck consists of a binary feature map with $L$ channels and 16 times smaller spatial resolution in both width and height.
Toderici~\etal use a stochastic binarization to allow a gradient signal through the bottleneck.
Mathematically, this reduces to \textit{REINFORCE}~\cite{w-ssgac-92} on sigmoidal activations.
At inference time, the most likely state is selected.

This architecture yields state-of-the-art image compression performance.
However, it fails to exploit any temporal redundancy.

\vspace{-2mm}
\paragraph{Video compression.}
Modern video codecs process I-frames using an image encoder $E_I$ and decoder $D_I$.
P-frames store a block motion estimate $\mv \in \RR^{W\times H\times 2}$, similar to an optical flow field, and a residual image $\res$, capturing the appearance changes not explained by motion.
Both motion estimate and residual are jointly compressed using entropy coding.
The original color frame is then recovered by
\begin{align}
I_i^{(t)} = I^{(t-1)}_{i - \mv^{(t)}_i} + \res^{(t)}_i, \label{eq:motion_compensation}
\end{align}
for every pixel $i$ in the image.
The compression is uniquely defined by a block structure and motion estimate $\mv$. The residual is simply the difference between the motion interpolated image and the original.

In this paper, we propose a more general view on video compression through image interpolation.
We augment image interpolation network with motion information and add a compressible bottleneck layer.
\begin{figure*}[t]
  \begin{subfigure}[t]{0.38\textwidth}
    \center
    \includegraphics[height=4cm, page=1]{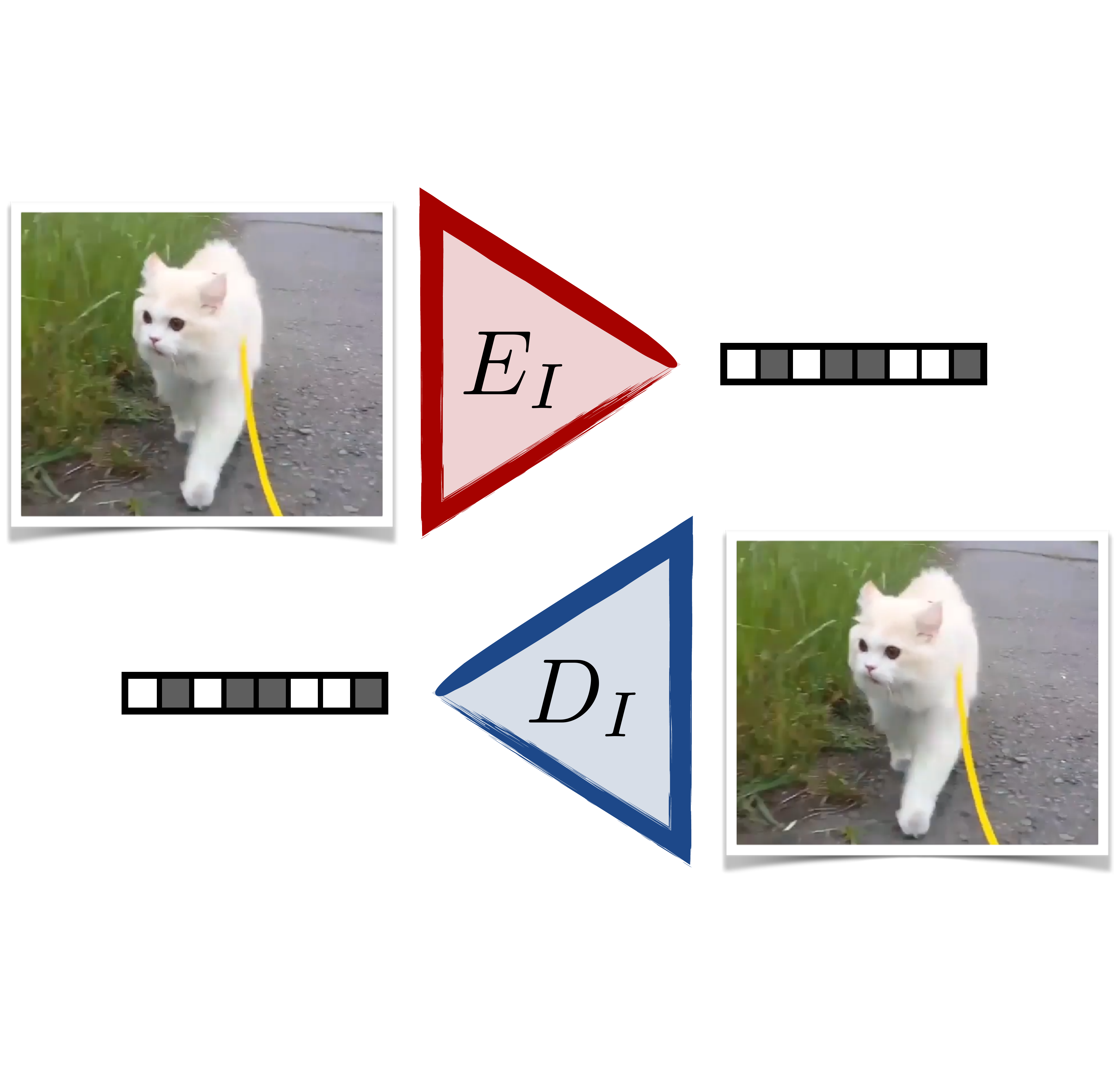}%
    \caption{I-frame model.}
    \lblfig{frame_model}
  \end{subfigure}\hfill
  \begin{subfigure}[t]{0.61\textwidth}
    \center
    \includegraphics[height=3.8cm,page=1]{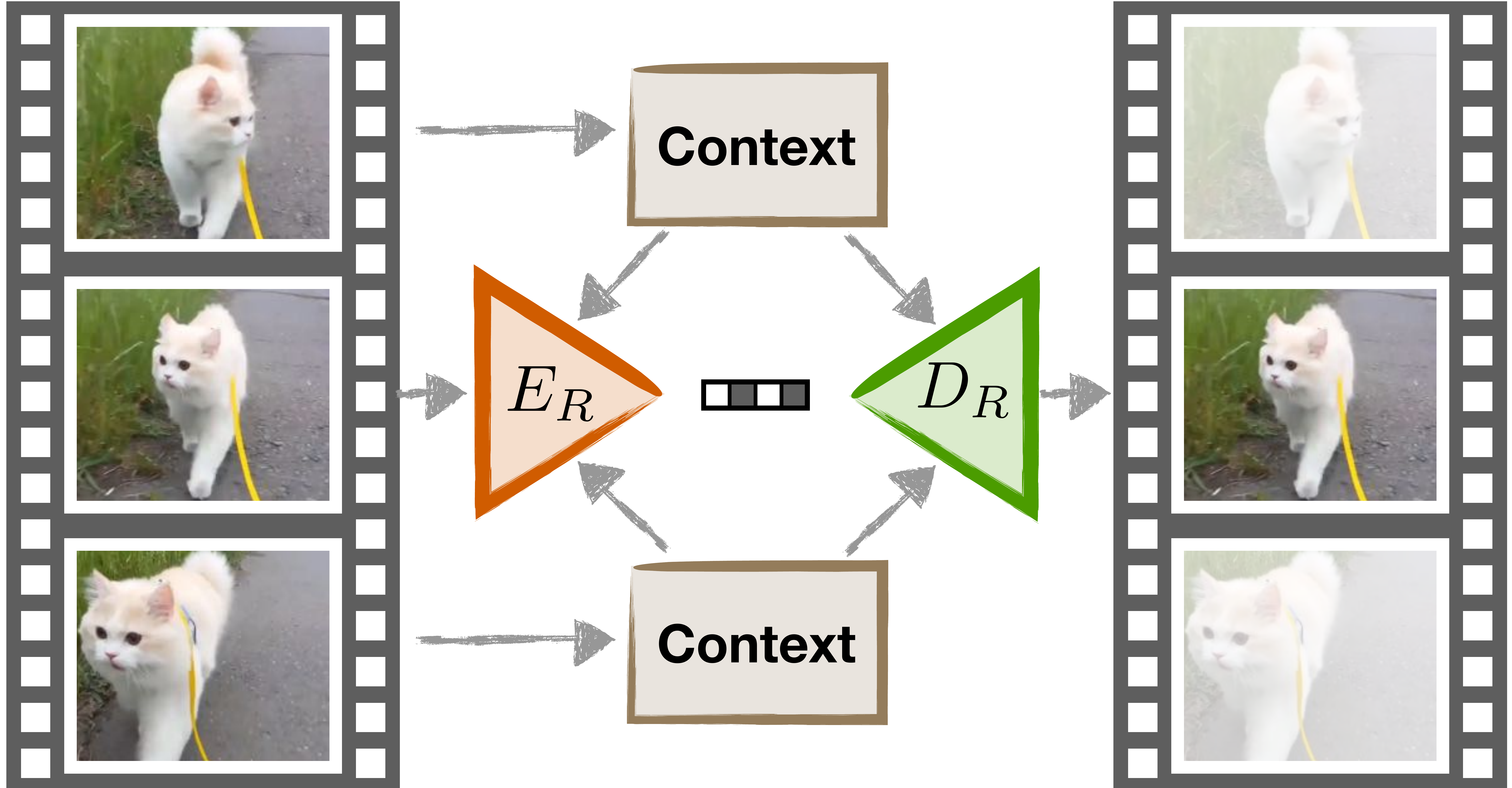}%
    \caption{Final interpolation model.}
    \lblfig{interpol_model}
  \end{subfigure}
  \vspace{-2mm}
  \caption{Our model is composed of an image compression model that compresses the key frames,
  and a conditional interpolation model that interpolates the remaining frames.}
  \lblfig{model}
\end{figure*}

\section{Video Compression through Interpolation}\label{sec:model}

Our codec first encodes I-frames using the compression algorithm of Toderici~\etal,
see \reffig{frame_model}.
We chose every $n$-th frame as an I-frame.
The remaining $n-1$ frames are interpolated.
We call those frames R-frames, as they reference other frames.
We choose $n=12$ in practice, but also experimented with larger groups of pictures.
We will first discuss our basic interpolation network, and then show a hierarchical interpolation setup, that further reduces the bitrate.

\subsection{Interpolation network}
In the simplest version of our codec, all R-frames use a blind interpolation network to interpolate between two key-frames $I_1$ and $I_2$.
Specifically, we train a context network $C: I \to \{f^{(1)},f^{(2)},\ldots\}$ to extract a series of feature maps $f^{(l)}$ of various spatial resolutions.
For notational simplicity let $f := \{f^{(1)},f^{(2)},\ldots\}$ be the collection of all context features.
In our implementation, we use the upconvolutional feature maps of a U-net architecture with increasing spatial resolution $\frac{W}{8}\times\frac{H}{8}$, $\frac{W}{4}\times\frac{H}{4}$, $\frac{W}{2}\times\frac{H}{2}$, $W\times H$, in addition to the original image.

We extract context features $f_1$ and $f_2$ for key-frames $I_1$ and $I_2$ respectively, and train a network $D$ to interpolate the frame $\hat{I} := D\rbr{f_1, f_2}$.
$C$ and $D$ are trained jointly.
This simple model favors a high compression rate over image quality, as none of the R-frames capture any information not present in the I-frames.

Without any further information, it is impossible for the network to faithfully reconstruct a frame.
What can we provide to the network to make interpolation easier?

\vspace{-2mm}
\paragraph{Motion compensated interpolation.}
A great candidate is ground truth motion information.
It defines where pixels move 
through time and greatly disambiguates interpolation.
We tried both optical flow~\cite{farneback2003two} and block motion estimation~\cite{richardson2002video}.
Block motion estimates are easier to compress, but optical flow retains finer details.

We use the motion information to warp each context feature map
\begin{align}
\tilde f^{(l)}_i = f^{(l)}_{i - \mv_i},
\end{align}
at every spatial location $i$.
We scale the motion estimation with the resolution of the feature map, and use bilinear interpolation for fractional locations.
The decoder now uses the warped context features $\tilde f$ instead, which allows it to focus solely on image creation, and ignore motion estimation.

Motion compensation greatly improves the interpolation network, as we will show in \secref{exp}.
However, it still only produces content seen in either reference image.
Variations beyond motion, such as change in lighting, deformation,
occlusion, etc.\ are not captured by this model.

Our goal is to encode the remaining information in a highly compact from.

\vspace{-2mm}
\paragraph{Residual motion compensated interpolation.}
Our final interpolation model combines motion compensated interpolation with a compressed residual information, capturing both the motion and appearance difference in the interpolated frames.
\figref{interpol_model} show an overview of the model.

We jointly train an encoder $E_R$, context model $C$ and interpolation network $D_R$.
The encoder sees the same information as the interpolation network, which allows it to compress just the missing information, and avoid a redundant encoding.
Formally, we follow the progressive compression framework of Toderici~\etal~\cite{toderici2017full}, and train a variable bitrate encoder and decoder conditioned on the warped context $\tilde f$:
\begin{align*}
  r_0 &:= I\\
  b_k &:= E_R(r_{k-1}, \tilde f_1, \tilde f_2, g_{k-1}),  & r_k &:= r_{k-1} - D_R(b_k, \tilde f_1, \tilde f_2, h_{k-1}),  & \text{for }k=1,2,\ldots
\end{align*}

This framework allows the encoder to learn a variable rate compression at high reconstruction quality.
The interpolation network generally requires fewer bits to encode temporally close images and more bits for images that are farther apart.
In one extreme, when key frames do not provide any meaningful signal to the interpolated frame, our residual motion compensated interpolation reduces to image compression.
In the other extreme, when the image content does not change, our algorithm reduces to a vanilla interpolation, and requires close to zero bits.

In the next section, we use this to our advantage, and design a hierarchical interpolation scheme, maximizing the number of temporally close interpolations.

\begin{figure}[b]
    \center
    \includegraphics[width=\linewidth,page=1]{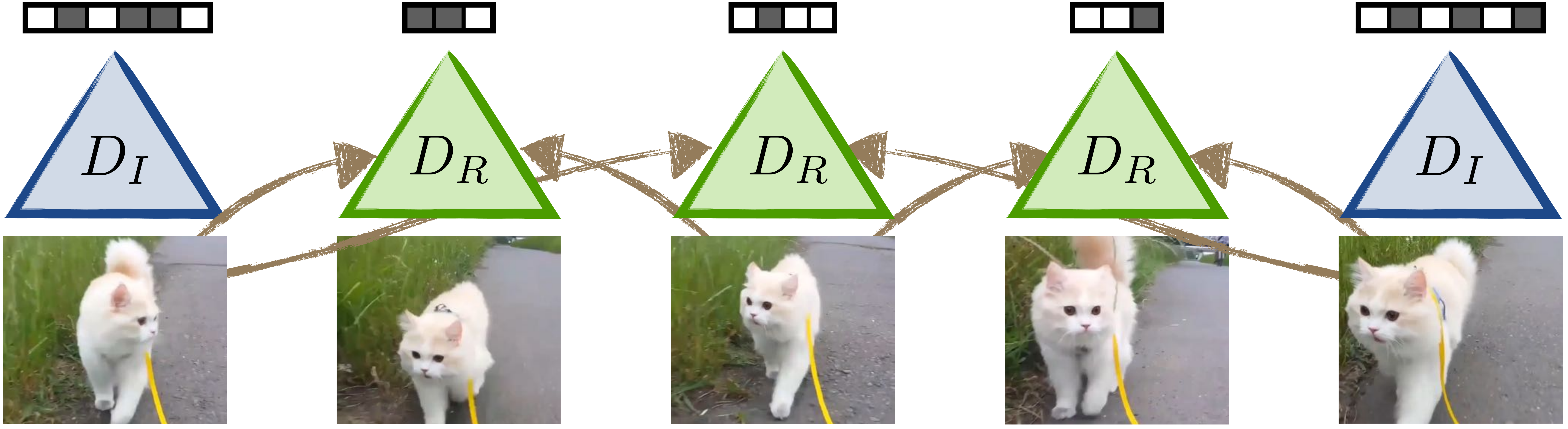}%
  \vspace{-2mm}
  \caption{ We apply the interpolation in a hierarchical manner. 
    Each level in the hierarchy uses previously decompressed images.}
  \label{fig:hierarchy}
\end{figure}

\subsection{Hierarchical interpolation}
The basic idea of hierarchical interpolation is simple: We interpolate some frames first, and use them as key-frames for the next level of interpolations.
See \reffig{hierarchy} for example.
Each interpolation model $\Mcal_{a,b}$ references $a$ frames into the past and $b$ frames into the future.
There are a few things we need to trade off.
First, every level in our hierarchical interpolation compounds error.
The shallower the hierarchy, the fewer errors compound.
In practice, the error propagation for more than three levels in the hierarchy significantly reduces the performance of our codec.
Second, we need to train a different interpolation network $\Mcal_{a,b}$ for each temporal offset $(a, b)$, as different interpolations behave differently.
To maximally use each trained model, we repeat the same temporal offsets as often as possible.
Third, we need to minimize the sum of temporal offsets used in interpolation.
The compression rate directly relates to the temporal offset, hence minimizing the temporal offset reduces the bitrate.

Considering just the bitrate and the number of interpolation networks, the optimal hierarchy is a binary tree cutting the interpolation range in two at each level.
However, this cannot interpolate more than $n=2^3=8$ consecutive frames, without significant error propagation.
We extend this binary structure to $n=12$ frames, by interpolating at a spacing of three frames in the last level of the hierarchy.
For a sequence of four frames $I_1,\ldots,I_4$, we train an interpolation model $\Mcal_{1,2}$ that predicts frame $I_2$, given $I_1$ and $I_4$.
We use the exact same model $\Mcal_{1,2}$ to predict $I_3$, but flip the conditioned images $I_4$ and $I_1$.
This yields an equivalent model $\Mcal_{2,1}$ predicting the third instead of the second image in the series.
Combining this with an interpolation model $\Mcal_{3,3}$ and $\Mcal_{6,6}$ in a hierarchy, we extend the interpolation range from $n=8$ frames to $n=12$ frames while keeping the same number of models and levels.
We tried applying the same trick to all levels in the hierarchy, extending the interpolation to $n=27$ frames, but performance dropped, as we had more distant interpolations.

To apply this to a full video of $N$ frames, we divide them into $\ceil{N/n}$
groups of pictures (GOPs).
Two consecutive groups share the same boundary I-frame.
We apply our hierarchical interpolation to each group independently.

\vspace{-2mm}
\paragraph{Bitrate optimization.}
Each interpolation model at a level $l$ of the hierarchy, can choose to spend $K_l$ bits to encode an image.
Our goal is to minimize the overall bitrate, while maintaining a low distortion for all encoded frames.
The challenge here is that each selection of $K_l$ affects all lower levels, as errors propagate.
Selecting a globally optimal set of $\cbr{K_l}$ thus requires iterating through all possible combinations, which is infeasible in practice.

We instead propose a heuristic bitrate selection based on beam search.
For each level we chose from $m$ different bitrates.
We start by enumerating all $m$ possibilities for the I-frame model.
Next, we expand the first interpolation model with all $m$ possible bitrates, leading to $m^2$ combinations.
Out of these combinations, not all lead to a good MS-SSIM per bitrate, and we discard combinations not on the envelope of the MS-SSIM vs bitrate curve.
In practice, only $O(m)$ combinations remain.
We repeat this procedure for all levels of the hierarchy.
This reduces the search space from $m^L$ to $O(Lm^2)$ for an $L$-level hierarchy.
In practice, this yields sufficiently good bitrates.

\subsection{Implementation}

\paragraph{Architecture.}
Our encoder and decoder (interpolation network) architecture follows the image compression model in Toderici \etal~\cite{toderici2017full}.
While Toderici \etal use $L=32$ latent bits to compress an image, 
we found that for interpolation, $L=8$ bits suffice for distance $3$ and $L=16$ for distance $6$ and $12$.
This yields a bitrate of $0.0625$ bits per pixel (BPP) and $0.03125$ BPP for each iteration respectively.

We use the original U-net \cite{ronneberger2015u} as the context model.
To speed-up training and save memory, we reduce the number of channels of all filters by half.
We did not observe any significant performance degradation.

To make it compatible with our architecture, 
we remove the final output layer and takes the feature maps at the resolutions
that are $2\times$, $4\times$, $8\times$ smaller than the original input image.

\paragraph{Conditional encoder and decoder.}
To add the information of the context frames into the encoder and decoder, we fuse the U-net features with the individual Conv-LSTM layers.
Specifically, we perform the fusion before each Conv-LSTM layer by concatenating the corresponding U-net features of the same spatial resolution.
To increase the computational efficiency, we selectively turn some of the conditioning off in both encoder and decoder.
This was tuned for each interpolation network; see supplementary material for details.

To help the model compare context frames and the target frame side-by-side,
we additionally stack the two context frames with target frame,
resulting in a $9$-channel image, and use that instead as the encoder input.

\paragraph{Entropy coding.}
Since the model is fully-convolutional, 
it uses the same number of bits for all locations of an image. 
This disregards the fact that information is not distributed uniformly in an image.
Following Mentzer \etal~\cite{mentzer2018conditional}, we train a 3D Pixel-CNN on the
$\cbr{0,1}^{\nicefrac{W}{16} \times \nicefrac{H}{16} \times L}$ binary representations
to obtain the probability of each bit sequentially.
We then use this probability with adaptive arithmetic coding to encode the feature map.
See supplementary material for more details. 

\paragraph{Motion compression.}
We store forward and backward block motion estimates as a lossless 4-channel WebP~\cite{webp} image.
For optical flow we train a separate lossy deep compression model, as lossless WebP was unable to compress the flow field.

\section{Experiments}
\lblsec{exp}
In this section, we perform a detailed analysis on the series of interpolation models (\secref{ablation}), 
and present both quantitative and qualitative (\secref{qual})
evaluation of our approach.

\vspace{-2mm}
\paragraph{Datasets and Protocol.}
We train our models using videos from the Kinetics dataset \cite{carreira2017quo}.
We only use videos with a width and height greater than $720$px.
To remove artifacts induced by previous compression, we downsample those high resolution videos to $352\times288$px.
We allow the aspect ratio to change.
The resulting dataset contains 2.8M frames in 75K videos.
We train on 65K, use 5K for validation, and 5k for testing.
For faster testing on Kinetics, we only use a single group of $n=12$ pictures per video.

We additionally test our method on two raw video datasets, Video Trace Library (VTL) \cite{vtl} 
and Ultra Video Group (UVG) \cite{uvg}. 
The VTL dataset contains $\sim40$K frames of resolution $352\times288$ in 20 videos. 
The UVG dataset contains $3,900$ frames of resolution $1920\times1080$ in 7 videos.

We evaluate our method based on the compression rate in bits per pixel (BPP),
and the quality of compression in multi-scale structural similarity (MS-SSIM)~\cite{wang2003multiscale}
and peak signal-to-noise ratio (PSNR).
We report the average performance of all videos,
as opposed to the average of all frames, as our final performance.
We use a GOP size of $n=12$ frames, for all algorithms unless otherwise stated.

\vspace{-2mm}
\paragraph{Training Details.}
All of our models are trained from scratch for 200K iterations using ADAM \cite{kingma2014adam},
with gradient norms clipped at $0.5$.
We use a batch size of 32 and a learning rate of 0.0005, 
which is divided by 2 when the validation MS-SSIM plateaus.
We augment the data through horizontal flipping.
For image models we train on $96\times96$ random crops, and for
the interpolation models we train on $64\times64$ random crops.
We train all models with $10$ reconstruction iterations.

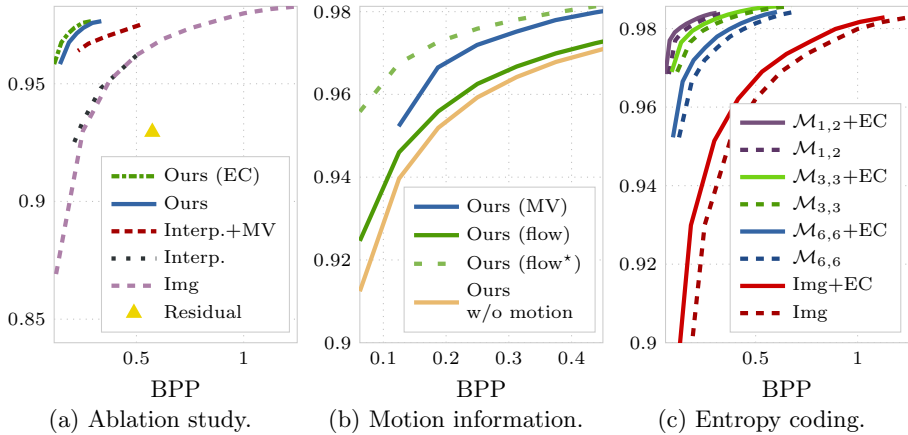
\begin{figure}[t]
\center
\begin{subfigure}[b]{0.33\textwidth}
      \begin{tikzpicture}
      \begin{axis}[
          legend pos=south east,
          width=1.2\linewidth,
          height=6cm,
          xlabel={BPP},
          ymin=0.84,
          every axis plot/.append style={ultra thick},
          cycle multiindex* list={
            mycolorlist
                \nextlist
            mystylelist
          },
      ]
      \pgfplotsset{every tick label/.append style={font=\scriptsize}}
      \addplot
          table {plotdata/trace_ours_ec_ssim.data};
          \addlegendentry{Ours (EC)}
      \addplot
          table {plotdata/trace_ours_ssim.data};
          \addlegendentry{Ours}
      \addplot
          table {plotdata/trace_interpolate_warp.data};
          \addlegendentry{Interp.+MV}
      \addplot
          table {plotdata/trace_interpolate.data};
          \addlegendentry{Interp.}
      \addplot
          table {plotdata/trace_i_ssim.data};
          \addlegendentry{Img}
      \addplot[only marks, butter2, mark=triangle*]
          table {plotdata/trace_residual_ssim.data};
          \addlegendentry{Residual}
      \end{axis}
      \end{tikzpicture}\vspace{-2mm}
      \caption{Ablation study.}\label{fig:ablation}
\end{subfigure}\hfill
\begin{subfigure}[b]{0.33\textwidth}
      \begin{tikzpicture}
      \begin{axis}[
          legend pos=south east,
          width=1.2\linewidth,
          height=6cm,
          xlabel={BPP},
          xmax=0.45,
          ymin=0.9,
          every axis plot/.append style={ultra thick},
          cycle multiindex* list={
            mycolorlistdark
                \nextlist
            mystylelist
          },
      ]
      \pgfplotsset{every tick label/.append style={font=\scriptsize}}
      \addplot[skyblue2, solid]
          table {plotdata/trace_d66_ssim.data};
          \addlegendentry{Ours (MV)}
      \addplot[chameleon3, solid]
          table {plotdata/dist6_flow_warp.data};
          \addlegendentry{Ours (flow)}
      \addplot[chameleon3!70, loosely dashed]
          table {plotdata/dist6_trace_perfect_flow_ssim.data};
          \addlegendentry{Ours (flow$^\star$)}
      \addplot[chocolate1]
          table {plotdata/dist6_no_warp.data};
          \addlegendentry{Ours\\w/o motion}
      \end{axis}
      \end{tikzpicture}\vspace{-2mm}
      \caption{Motion information.
      }\label{fig:mv}
\end{subfigure}\hfill
\begin{subfigure}[b]{0.33\textwidth}
      \begin{tikzpicture}
      \begin{axis}[
          legend pos=south east,
          width=1.2\linewidth,
          height=6cm,
          xlabel={BPP},
          ymin=0.9,
          every axis plot/.append style={ultra thick},
      ]
      \pgfplotsset{every tick label/.append style={font=\scriptsize}}
      \addplot[plum2]
          table {plotdata/trace_d12_ssim_ec.data};
          \addlegendentry{$\Mcal_{1,2}$+EC}
      \addplot[dashed, plum3]
          table {plotdata/trace_d12_ssim.data};
          \addlegendentry{$\Mcal_{1,2}$}
      \addplot[chameleon2]
          table {plotdata/trace_d33_ssim_ec.data};
          \addlegendentry{$\Mcal_{3,3}$+EC}
      \addplot[dashed, chameleon3]
          table {plotdata/trace_d33_ssim.data};
          \addlegendentry{$\Mcal_{3,3}$}
      \addplot[skyblue2]
          table {plotdata/trace_d66_ssim_ec.data};
          \addlegendentry{$\Mcal_{6,6}$+EC}
      \addplot[dashed, skyblue3]
          table {plotdata/trace_d66_ssim.data};
          \addlegendentry{$\Mcal_{6,6}$}
      \addplot[scarletred2]
          table {plotdata/trace_i_ssim_ec.data};
          \addlegendentry{Img+EC}
      \addplot[dashed, scarletred3]
          table {plotdata/trace_i_ssim.data};
          \addlegendentry{Img}
      \end{axis}
      \end{tikzpicture}\vspace{-2mm}
      \caption{Entropy coding.}\label{fig:entropy}
\end{subfigure}%
\caption{MS-SSIM of different models evaluated on the VTL dataset.}
\end{figure}

\subsection{Ablation study}\label{sec:ablation}
We first evaluate the series of image interpolation models in \refsec{model} on the VTL dataset.
\reffig{ablation} shows the results.

We can see that image compression model requires by far the highest BPP to achieve high visual quality
and performs poorly in the low bitrate region.
This is not surprising as it does not exploit any temporal redundancy and needs
to encode everything from scratch.
Vanilla interpolation does not work much better.
We present results for interpolation from 1 to 4 frames, using the best image compression model.
While it exploits the temporal redundancy,
it fails to accurately reconstruct the image.

Motion-compensated interpolation works significantly better.
The additional motion information
disambiguates the interpolation, improving the accuracy.
The presented BPP includes the size of motion vectors. %

Our final model efficiently encodes residual information and makes good use of hierarchical referencing.
It achieves the best performance when combined with entropy coding.
Note the large performance gap between our method and the image compression model 
in the low bitrate regime -- our model effectively uses context information and achieves a good performance even with very few bits per pixel.

As a sanity check, we further implemented a simple deep codec that uses image compression to encode the residual $\Rcal$ in traditional codecs.
This simple baseline stores the video as the encoded residuals, compressed motion vectors, in addition to key frames compressed by a separate deep image compression model.
The residual model struggles to learn patterns from noisy residual images, and works worse than an image-only compression model.
This suggests that trivially extending deep image compression to videos is not sufficient.
Our end-to-end interpolation network performs considerably better.

\paragraph{Motion.}
Next, we analyze different motion estimation models, and compare optical flow to block motion vectors.
For optical flow, we use the OpenCV implementation of the Farneb{\"a}ck's algorithm \cite{farneback2003two}.
For motion compensation, we use the same algorithm as H.264. 

\figref{mv} shows the results of the $\Mcal_{6,6}$ model with both motion sources.
Using motion information clearly helps improve the performance of the model, despite the overhead of motion compression.
Block motion estimation (MV) works significantly better than the optical flow based model (flow).
Almost all of this performance gain comes from better compressible motion information.
The block motion estimates are smaller, easier to compress, and fit in a lossless compression scheme.

To understand whether the worse performance of optical flow is due to the errors in flow compression
or the property of the flow itself, 
we further measure the \emph{hypothetical} performance upper bound 
of an optical flow based model \emph{assuming} a lossless flow compression at no additional cost (flow$^\star$).
As shown in \figref{mv}, this upper bound performs better than motion vectors, leaving room for improvement through compressible optical flow estimation.
However, finding such a compressible flow estimate is beyond the scope of this paper.
In the rest of this section, we use block motion estimates in all our experiments.

\paragraph{Individual interpolation models and entropy coding.}
\reffig{entropy} shows the performance of different interpolation models with and without entropy coding.
For all models, entropy coding saves up to $52\%$ BPP, at low bitrate regions, and at least $10\%$ BPP, at high bitrate region.
More interestingly, the short time-frame interpolation is almost free, achieving the same visual quality as an image-based model at one or two orders of magnitude fewer bits per pixel.
This shows that most of our bitrate saving comes from the interpolation models at lower levels in the hierarchy.

\vspace{-2mm}
\subsection{Comparison to prior work}\label{sec:quant}
\begin{figure}[t]
\center
\begin{subfigure}[b]{0.5\textwidth}
\vspace{0pt}
      \begin{tikzpicture}
      \begin{axis}[
          width=1.1\linewidth,
          height=6cm,
          ymin=0.82,
          xmax=0.4,
          legend pos=south east,
          every axis plot/.append style={ultra thick},
          cycle multiindex* list={
            mycolorlist3
                \nextlist
            mystylelist3
          },
      ]
      \addplot
          table {plotdata/ultra_h265_ssim.data};
          \addlegendentry{HEVC}
      \addplot+[on layer=foreground,]
          table {plotdata/ultra_ours_ec_ssim.data};
          \addlegendentry{Ours (EC)}
      \addplot+[on layer=foreground,]
          table {plotdata/ultra_ours_ssim.data};
          \addlegendentry{Ours}
      \addplot
          table {plotdata/ultra_h264_ssim.data};
          \addlegendentry{H.264}
      \addplot
          table {plotdata/ultra_mpeg4_ssim.data};
          \addlegendentry{MPEG-4 Part 2}
      \addplot
          table {plotdata/ultra_i_ssim.data};
          \addlegendentry{Image}
      \end{axis}
      \end{tikzpicture}
\caption{MS-SSIM}
\end{subfigure}\hfill
\begin{subfigure}[b]{0.5\textwidth}
\vspace{0pt}
      \begin{tikzpicture}
      \begin{axis}[
          width=1.1\linewidth,
          height=6cm,
          ymin=30,
          xmax=0.4,
          legend pos=south east,
          every axis plot/.append style={ultra thick},
          cycle multiindex* list={
            mycolorlist3
                \nextlist
            mystylelist3
          },
      ]
      \addplot
          table {plotdata/ultra_h265_psnr.data};
          \addlegendentry{HEVC}
      \addplot+[on layer=foreground,]
          table {plotdata/ultra_ours_ec_psnr.data};
          \addlegendentry{Ours (EC)}
      \addplot+[on layer=foreground,]
          table {plotdata/ultra_ours_psnr.data};
          \addlegendentry{Ours}
      \addplot
          table {plotdata/ultra_h264_psnr.data};
          \addlegendentry{H.264}
      \addplot
          table {plotdata/ultra_mpeg4_psnr.data};
          \addlegendentry{MPEG-4 Part 2}
      \addplot
          table {plotdata/ultra_i_psnr.data};
          \addlegendentry{Image}
      \end{axis}
      \end{tikzpicture}
\caption{PSNR (dB)}
\end{subfigure}\vspace{-2mm}
\caption{Performance on the UVG dataset. }\label{fig:uvg}\vspace{2mm}
\begin{subfigure}[b]{0.5\textwidth}
\vspace{0pt}
      \begin{tikzpicture}
      \begin{axis}[
          width=1.1\linewidth,
          height=6cm,
          legend pos=south east,
          every axis plot/.append style={ultra thick},
          cycle multiindex* list={
            mycolorlist2
                \nextlist
            mystylelist2
          },
      ]
      \addplot
          table {plotdata/trace_h265_ssim.data};
          \addlegendentry{HEVC}
      \addplot+[on layer=foreground,]
          table {plotdata/trace_ours_ec_ssim.data};
          \addlegendentry{Ours (EC)}
      \addplot+[on layer=foreground,]
          table {plotdata/trace_ours_ssim.data};
          \addlegendentry{Ours}
      \addplot
          table {plotdata/trace_h264_ssim.data};
          \addlegendentry{H.264}
      \addplot
          table {plotdata/trace_mpeg4_ssim.data};
          \addlegendentry{MPEG-4 Part 2}
      \addplot
          table {plotdata/trace_h261_ssim.data};
          \addlegendentry{H.261}
      \addplot
          table {plotdata/trace_i_ssim.data};
          \addlegendentry{Image}
      \end{axis}
      \end{tikzpicture}
\caption{MS-SSIM}
\end{subfigure}\hfill%
\begin{subfigure}[b]{0.5\textwidth}
      \begin{tikzpicture}
      \begin{axis}[
          width=1.1\linewidth,
          height=6cm,
          legend pos=south east,
          every axis plot/.append style={ultra thick},
          cycle multiindex* list={
            mycolorlist2
                \nextlist
            mystylelist2
          },
      ]
      \addplot
          table {plotdata/trace_h265_psnr.data};
          \addlegendentry{HEVC}
      \addplot+[on layer=foreground,]
          table {plotdata/trace_ours_ec_psnr.data};
          \addlegendentry{Ours (EC)}
      \addplot+[on layer=foreground,]
          table {plotdata/trace_ours_psnr.data};
          \addlegendentry{Ours}
      \addplot
          table {plotdata/trace_h264_psnr.data};
          \addlegendentry{H.264}
      \addplot
          table {plotdata/trace_mpeg4_psnr.data};
          \addlegendentry{MPEG-4 Part 2}
      \addplot
          table {plotdata/trace_h261_psnr.data};
          \addlegendentry{H.261}
      \addplot
          table {plotdata/trace_i_psnr.data};
          \addlegendentry{Image}
      \end{axis}
      \end{tikzpicture}
\caption{PSNR (dB)}
\end{subfigure}\vspace{-2mm}
\caption{Performance on the VTL dataset. }\label{fig:tvl}
\center
\begin{subfigure}[b]{0.5\textwidth}
\vspace{0pt}
      \begin{tikzpicture}
      \begin{axis}[
          width=1.1\linewidth,
          height=6cm,
          xmax=0.6,
          ymin=0.91,
          legend pos=south east,
          every axis plot/.append style={ultra thick},
          cycle multiindex* list={
            mycolorlist2
                \nextlist
            mystylelist2
          },
      ]
      \addplot
          table {plotdata/kinetics_h265_ssim.data};
          \addlegendentry{HEVC}
      \addplot+[on layer=foreground,]
          table {plotdata/kinetics_ours_ec_ssim.data};
          \addlegendentry{Ours (EC)}
      \addplot+[on layer=foreground,]
          table {plotdata/kinetics_ours_ssim.data};
          \addlegendentry{Ours}
      \addplot
          table {plotdata/kinetics_h264_ssim.data};
          \addlegendentry{H.264}
      \addplot
          table {plotdata/kinetics_mpeg4_ssim.data};
          \addlegendentry{MPEG-4 Part 2}
      \addplot
          table {plotdata/kinetics_h261_ssim.data};
          \addlegendentry{H.261}
      \addplot
          table {plotdata/kinetics_i_ssim.data};
          \addlegendentry{Image}
      \end{axis}
      \end{tikzpicture}
\caption{MS-SSIM}
\end{subfigure}\hfill
\begin{subfigure}[b]{0.5\textwidth}
\vspace{0pt}
      \begin{tikzpicture}
      \begin{axis}[
          width=1.1\linewidth,
          height=6cm,
          xmax=0.6,
          legend pos=south east,
          every axis plot/.append style={ultra thick},
          cycle multiindex* list={
            mycolorlist2
                \nextlist
            mystylelist2
          },
      ]
      \addplot
          table {plotdata/kinetics_h265_psnr.data};
          \addlegendentry{HEVC}
      \addplot+[on layer=foreground,]
          table {plotdata/kinetics_ours_ec_psnr.data};
          \addlegendentry{Ours (EC)}
      \addplot+[on layer=foreground,]
          table {plotdata/kinetics_ours_psnr.data};
          \addlegendentry{Ours}
      \addplot
          table {plotdata/kinetics_h264_psnr.data};
          \addlegendentry{H.264}
      \addplot
          table {plotdata/kinetics_mpeg4_psnr.data};
          \addlegendentry{MPEG-4 Part 2}
      \addplot
          table {plotdata/kinetics_h261_psnr.data};
          \addlegendentry{H.261}
      \addplot
          table {plotdata/kinetics_i_psnr.data};
          \addlegendentry{Image}
      \end{axis}
      \end{tikzpicture}
\caption{PSNR (dB)}
\end{subfigure}\vspace{-2mm}
\caption{Performance on the Kinetics-5K dataset.}\label{fig:kinetics}
\end{figure}

We now quantitatively evaluate our method on all three datasets
and compare our method with 
today's prevailing codecs, HEVC (H.265), H.264, MPEG-4 Part 2, and H.261.
For consistent comparison, we use the same GOP size, 12, for H.264 and HEVC.  
We test H.261 on only VTL and Kinetics-5K,
since it does not support high-resolution ($1920\times1080$) videos of the UVG dataset. 

Figures \ref{fig:uvg}-\ref{fig:kinetics} present the results.
Despite its simplicity, our model greatly outperforms MPEG-4 Part 2 and H.261, performs on par with H.264, and close to state-of-the-art HEVC.
In particular, on the high-resolution UVG dataset, it outperforms H.264 by a good margin and matches
HEVC in terms of PSNR.

Our testing datasets are not just large in scale ($>$100K frames of $>$5K videos),
they also consist of videos in a wide range of sizes (from $352\times288$ to $1920\times1080$),
time (from 1990s for most VTL videos to 2018 for Kinetics), 
quality (from professional UVG to user uploaded Kinetics),
and contents (from scenes, animals, to the 400 human activities in Kinetics).
Our model, trained on only one dataset, works well on all of them.

\label{sec:qual}
\begin{figure}[t]
\setlength{\tabcolsep}{0.1em}
\center
  \begin{tabular}{@{}cccc@{}}
    Ground truth& MPEG-4 Part 2&H.264& Ours\\
\includegraphics[width=0.2463\textwidth]{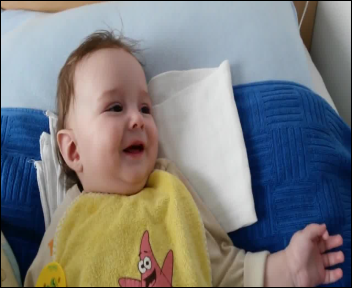}&
\includegraphics[width=0.2463\textwidth]{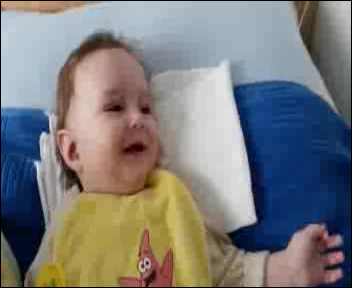}&
\includegraphics[width=0.2463\textwidth]{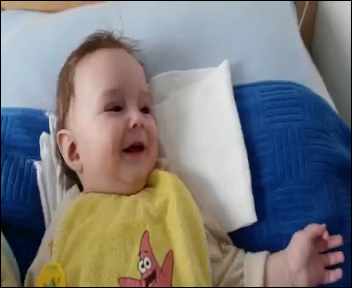}&
\includegraphics[width=0.2463\textwidth]{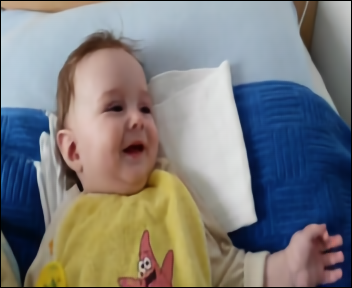}\\
\includegraphics[width=0.2463\textwidth]{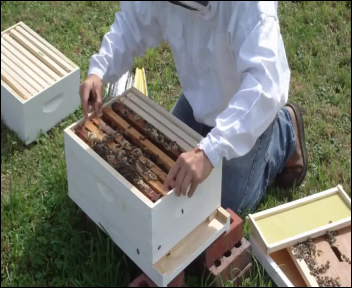}&
\includegraphics[width=0.2463\textwidth]{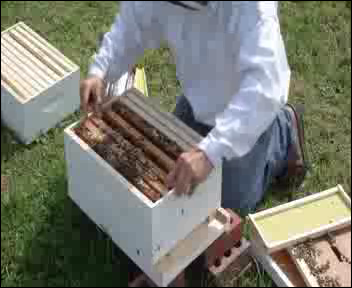}&
\includegraphics[width=0.2463\textwidth]{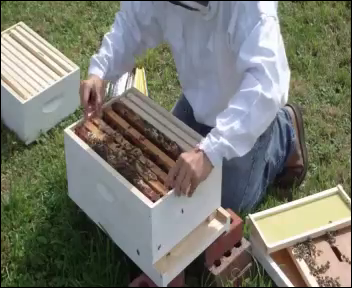}&
\includegraphics[width=0.2463\textwidth]{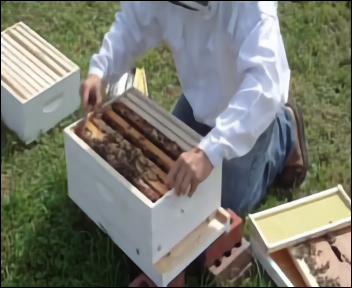}\\
    \multicolumn{4}{c}{(a) Kinetics-5K}\rule[-3mm]{0pt}{0pt}\\
\includegraphics[width=0.2463\textwidth]{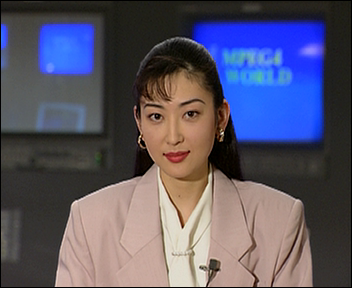}&
\includegraphics[width=0.2463\textwidth]{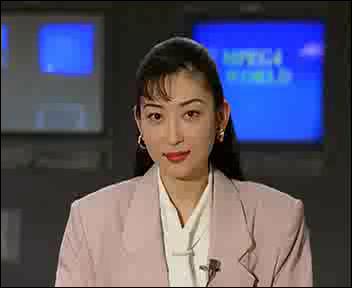}&
\includegraphics[width=0.2463\textwidth]{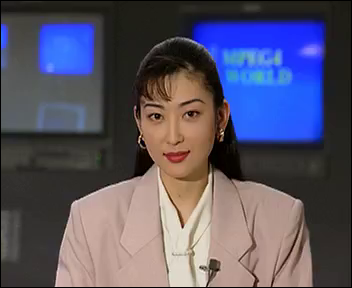}&
\includegraphics[width=0.2463\textwidth]{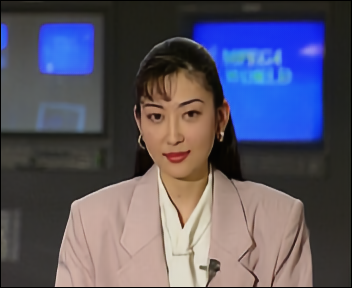}\\
\includegraphics[width=0.2463\textwidth]{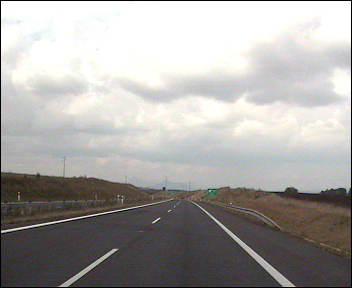}&
\includegraphics[width=0.2463\textwidth]{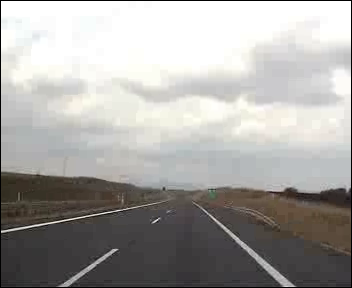}&
\includegraphics[width=0.2463\textwidth]{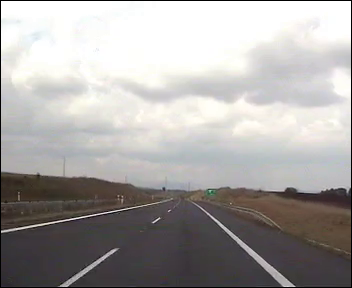}&
\includegraphics[width=0.2463\textwidth]{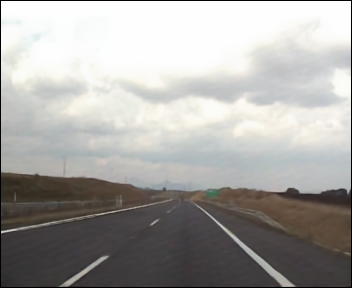}\\
    \multicolumn{4}{c}{(b) VTL }\rule[-3mm]{0pt}{0pt}\\
\includegraphics[width=0.2463\textwidth]{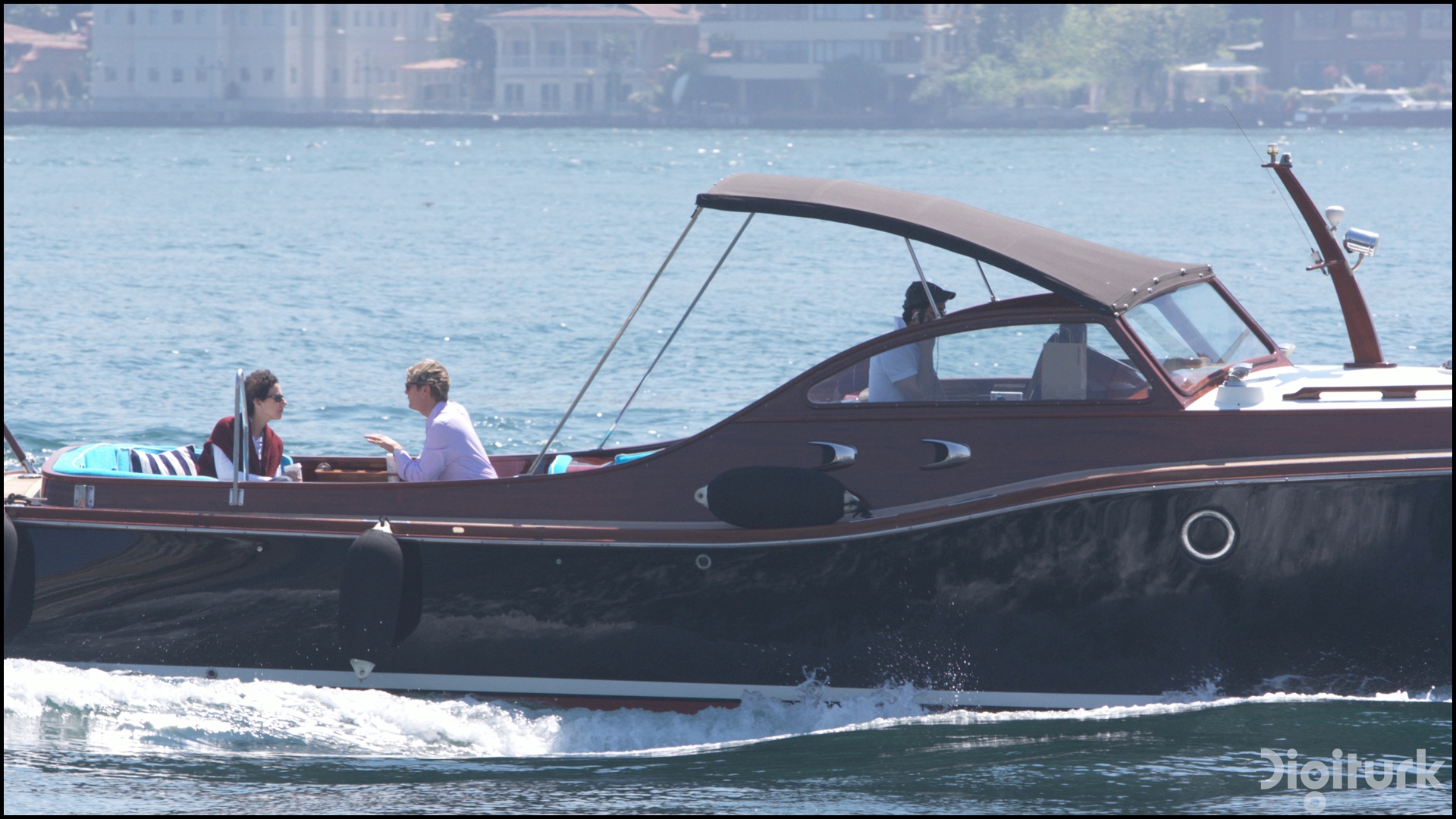}&
\includegraphics[width=0.2463\textwidth]{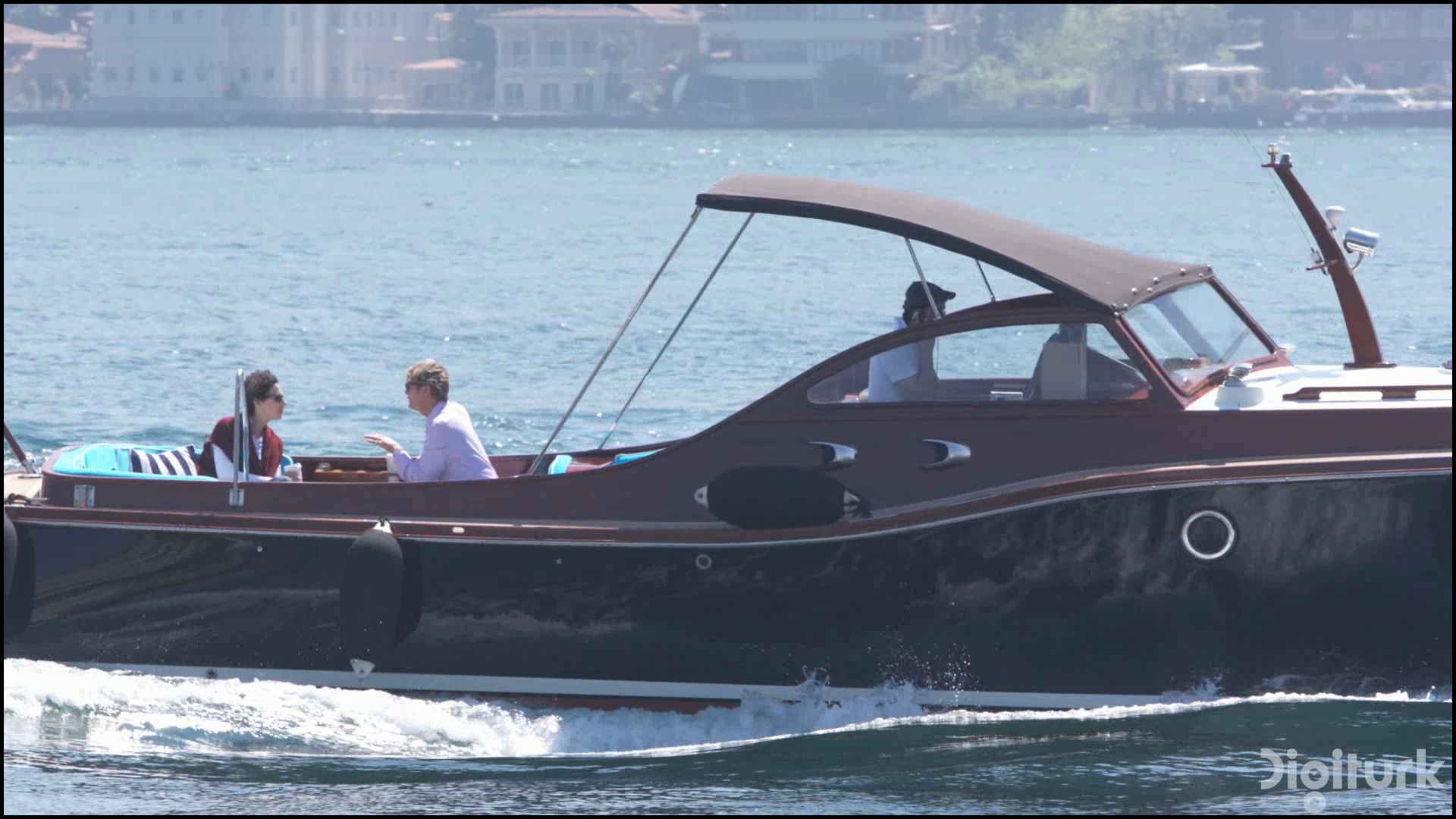}&
\includegraphics[width=0.2463\textwidth]{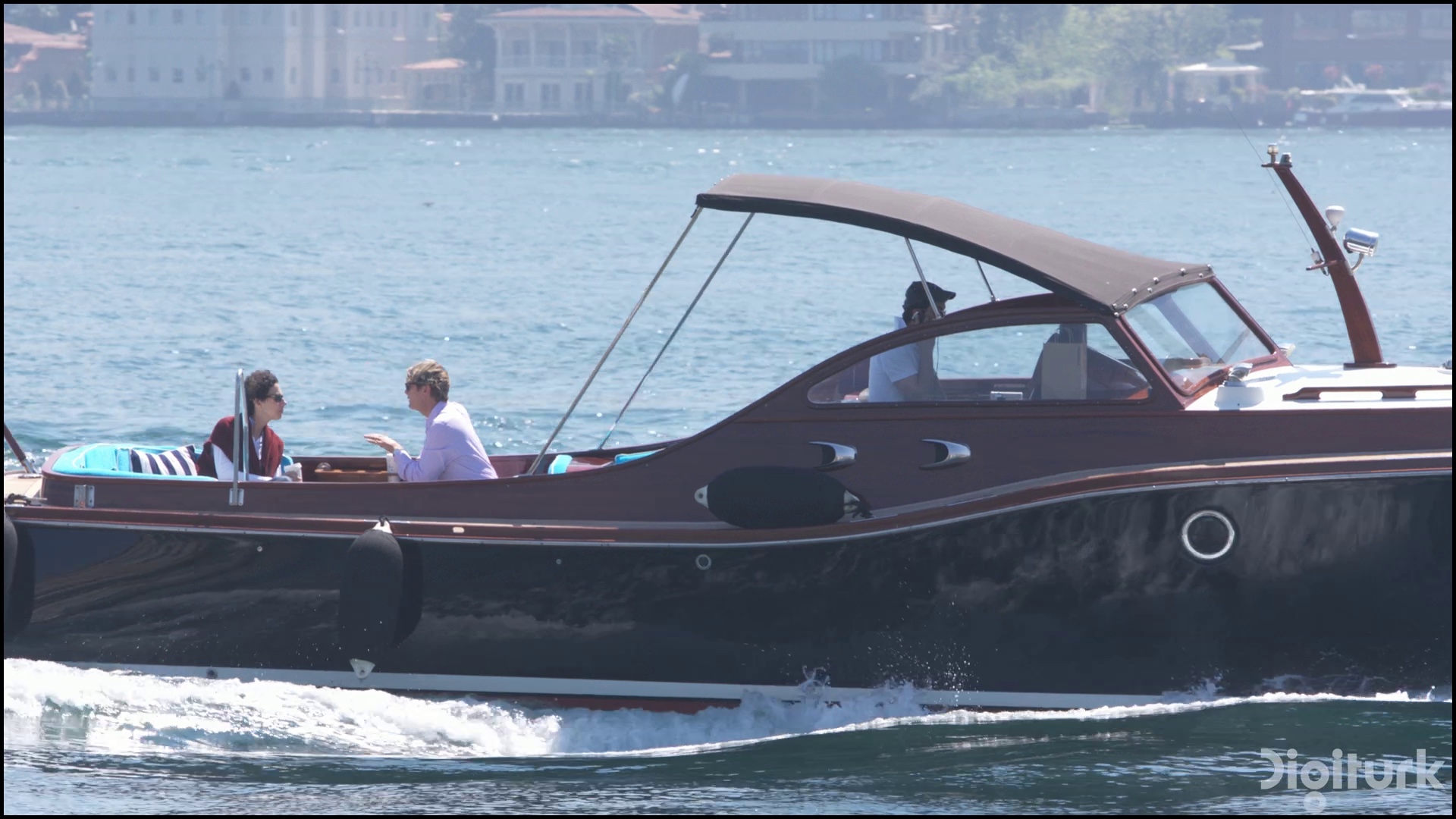}&
\includegraphics[width=0.2463\textwidth]{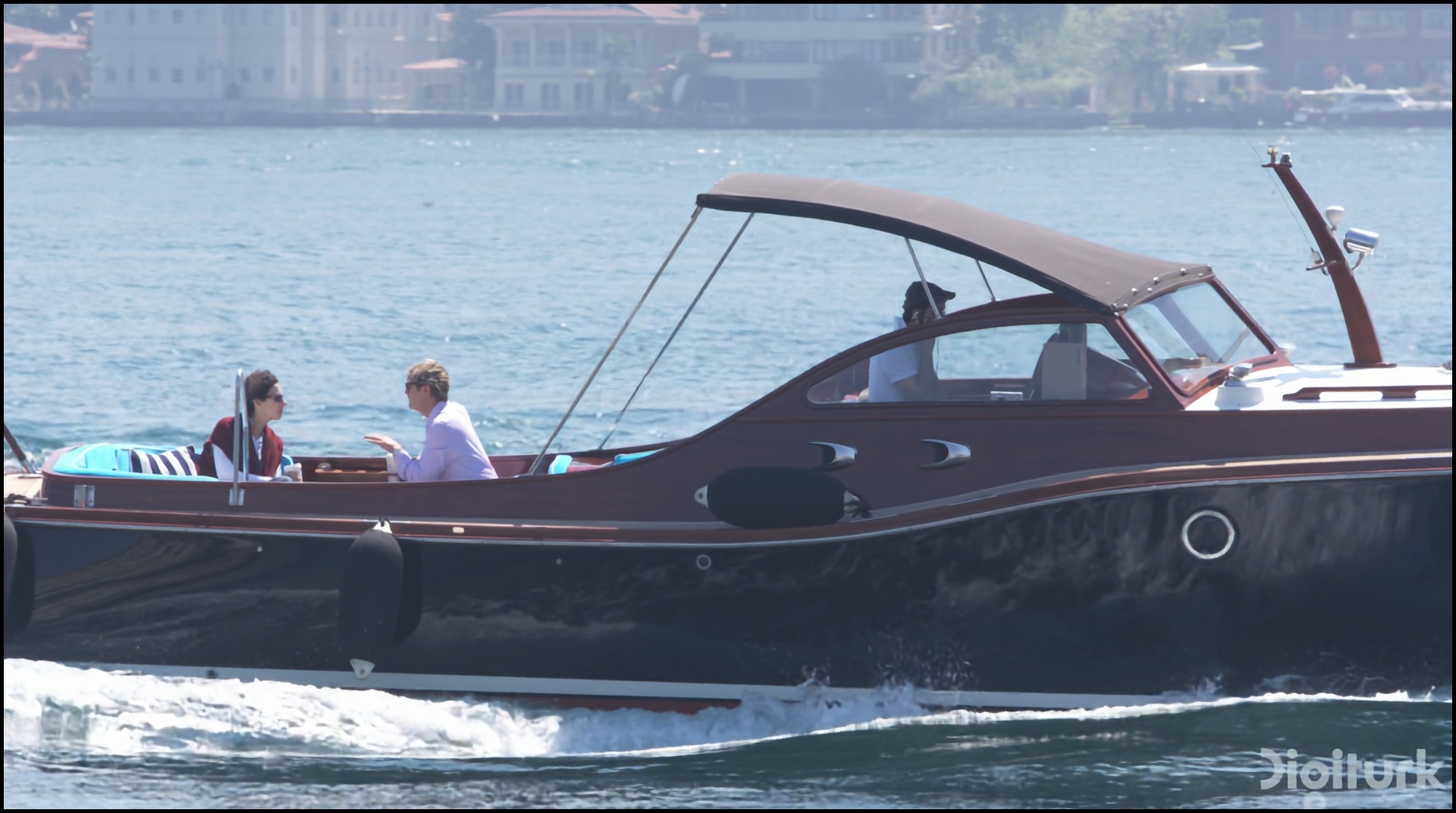}\\
\includegraphics[width=0.2463\textwidth]{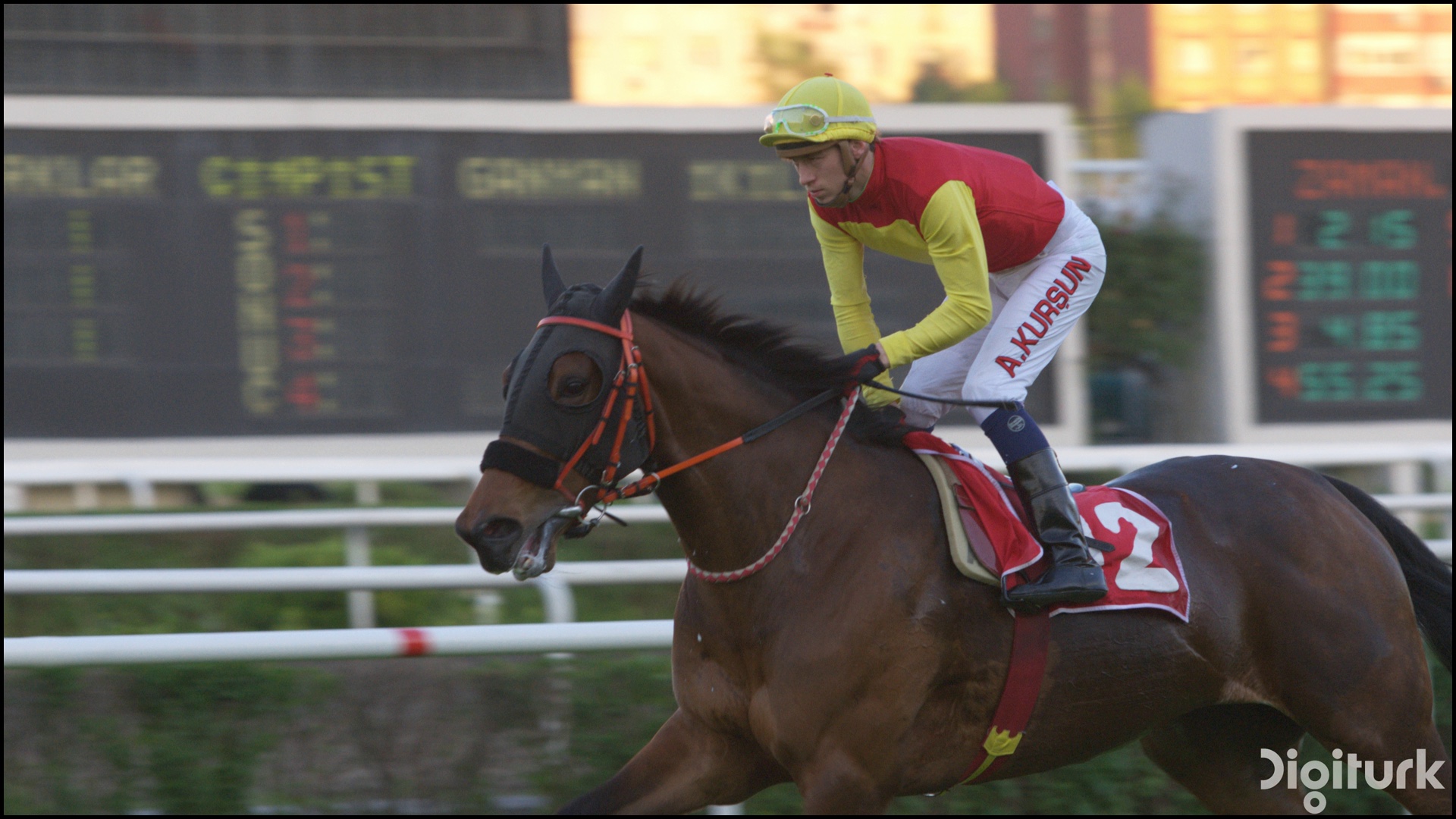}&
\includegraphics[width=0.2463\textwidth]{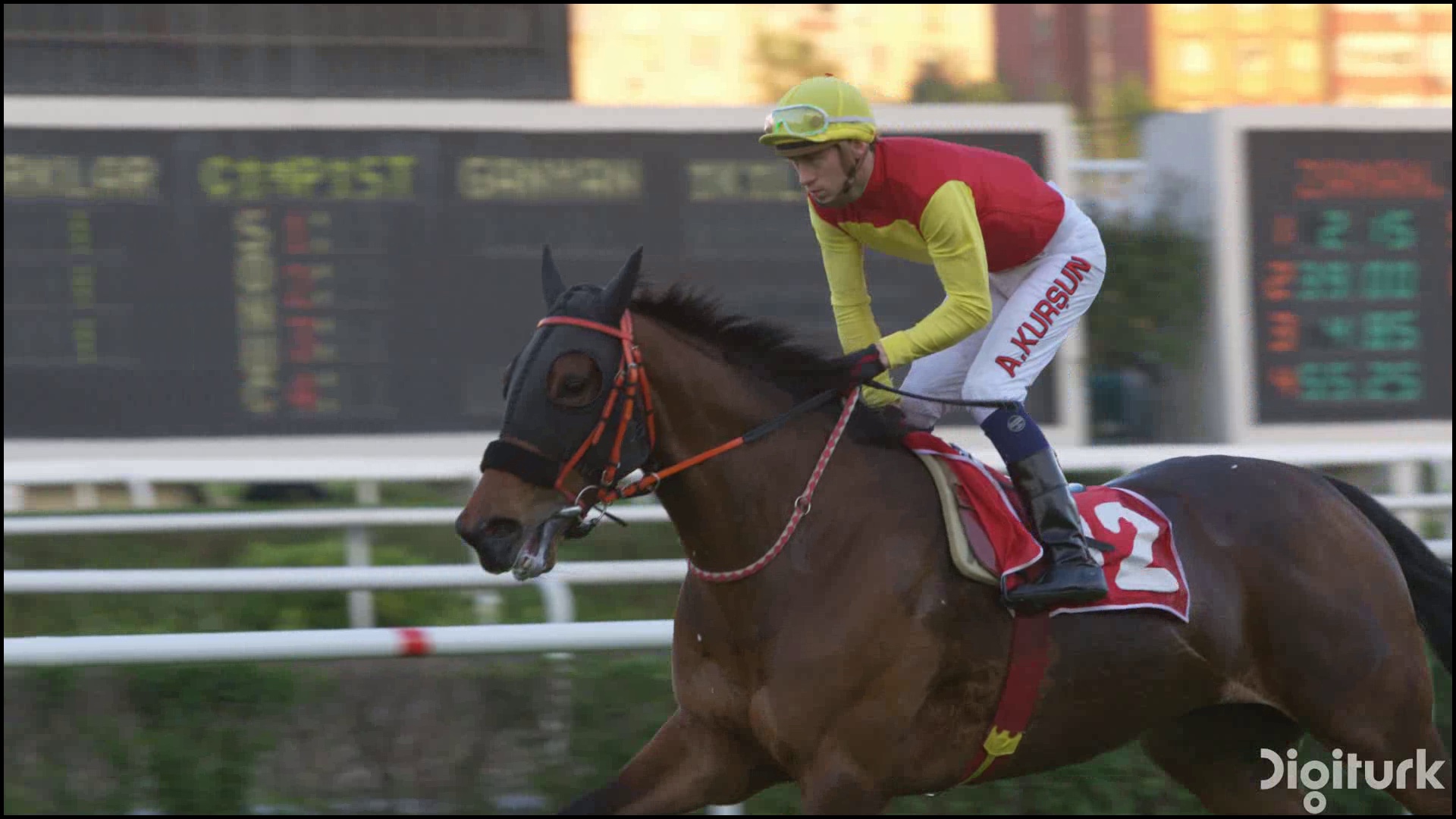}&
\includegraphics[width=0.2463\textwidth]{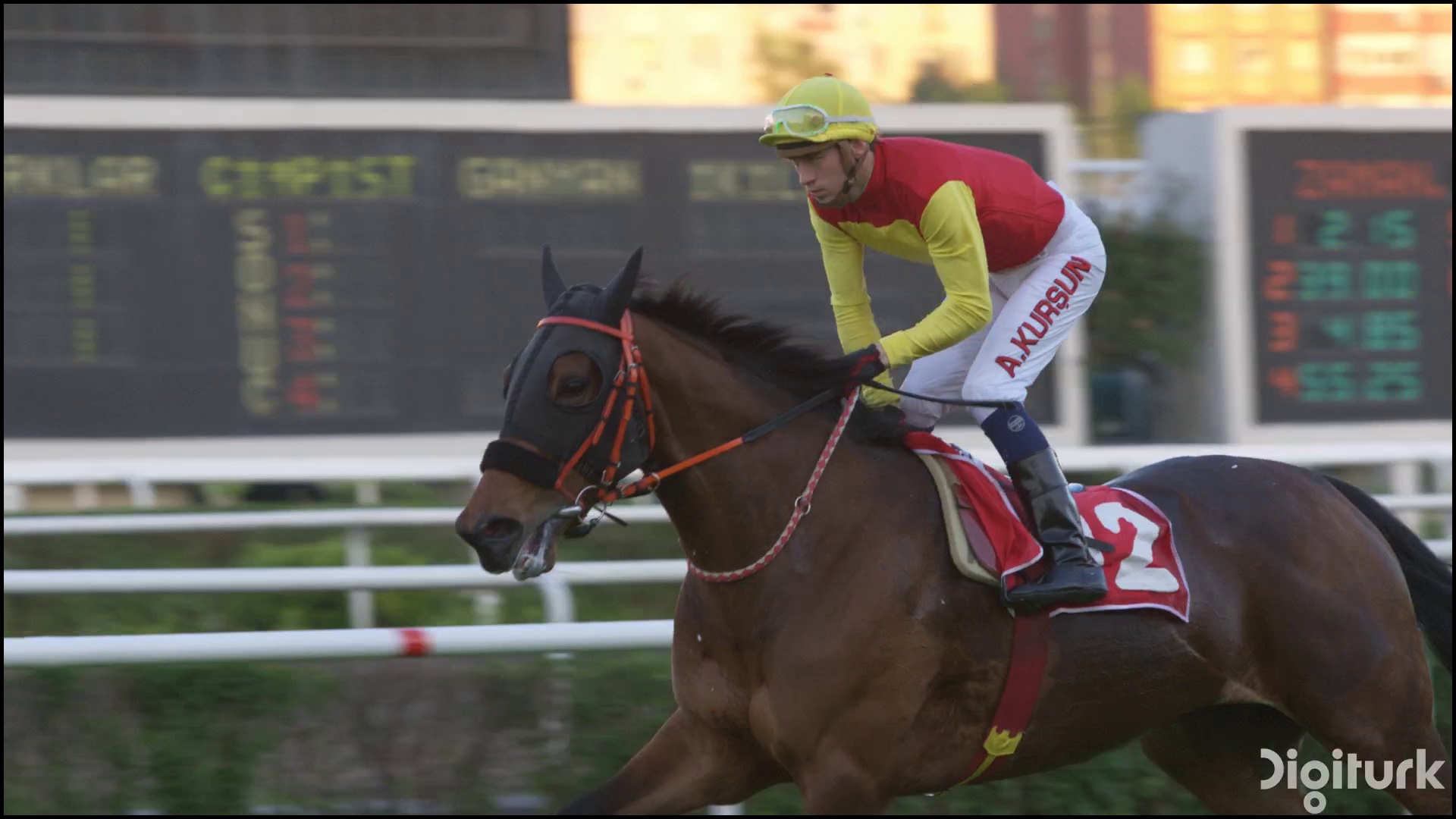}&
\includegraphics[width=0.2463\textwidth]{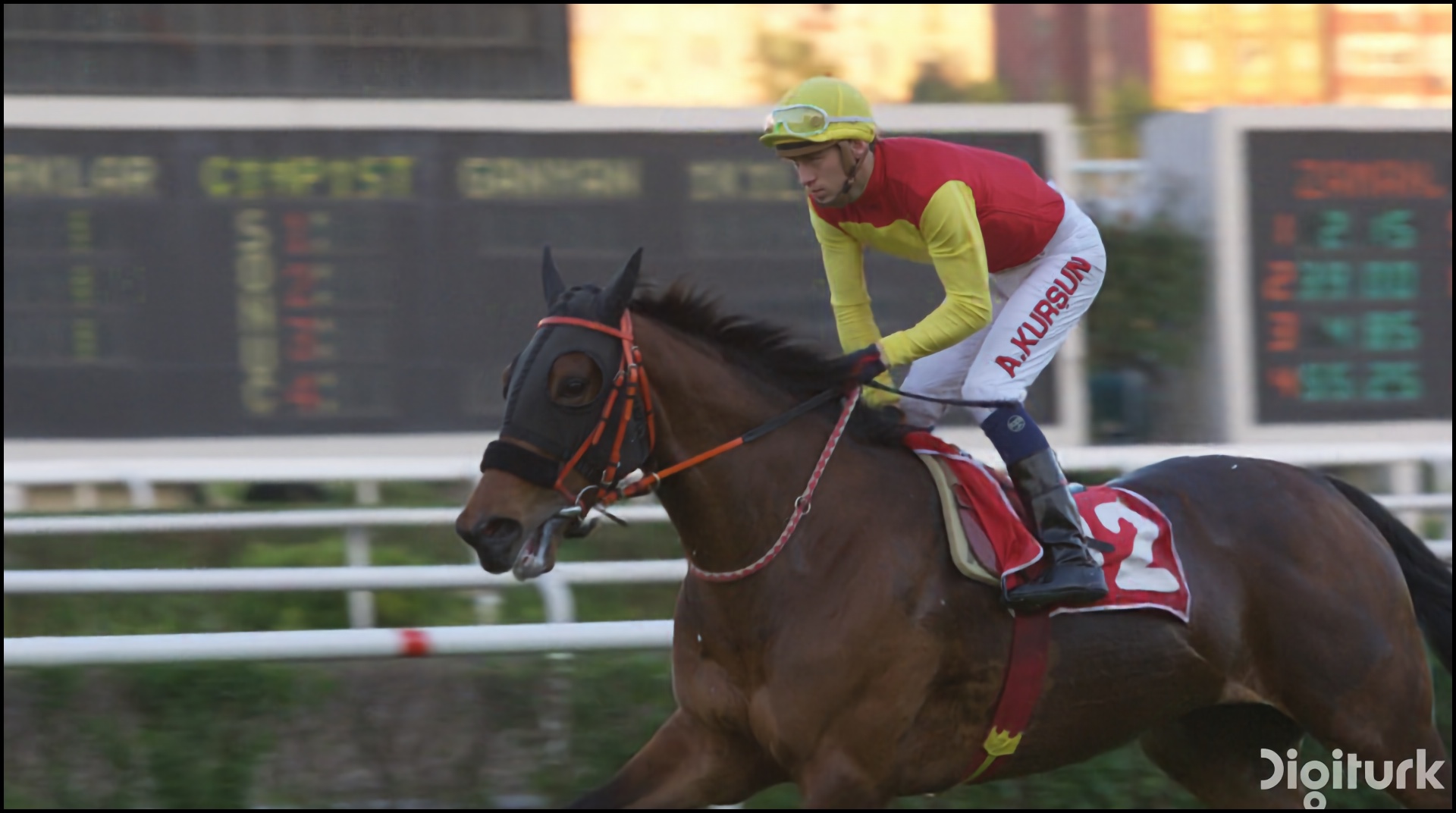}\\
    \multicolumn{4}{c}{(c) UVG }\rule[-3mm]{0pt}{0pt}\\
  \end{tabular}
  \vspace{-2mm}
  \caption{Comparison of compression results at $0.12\pm 0.01$ BPP.
  Our method shows faithful images without any blocky artifacts. (Best viewed on screen.)
  More examples and demo videos showing temporal coherence 
  are available at \url{https://chaoyuaw.github.io/vcii/}.}
  \label{fig:qual}
\end{figure}

Finally, we present qualitative results of 
three of the best performing models,
MPEG-4 Part 2, H.264 and ours in \figref{qual}.
All models here use $0.12\pm 0.01$ BPP.
We can see that in all datasets, our method shows faithful images without any blocky artifacts.
It greatly outperforms MPEG-4 Part 2 without bells and whistles, and matches state-of-the-art H.264. 

\vspace{-2mm}

\section{Conclusion}
This paper presents, to the best of our knowledge, the first end-to-end trained deep video codec.
It relies on repeated deep image interpolation.
To disambiguate the interpolation, we encode a few compressible bits of information representing information not inferred from the neighboring key frames.
This yields a faithful reconstruction instead of pure hallucination.
The network is directly trained to optimize reconstruction,
without prior engineering knowledge.

Our deep codec is simple, and
outperforms the prevailing codecs 
such as MPEG-4 Part 2 or H.261, matching state-of-the-art H.264.
We have not considered the engineering aspects such as
runtime or real-time compression.
We think they are important directions for future research.

In short, video compression powered by deep image interpolation
achieves state-of-the-art performance
without sophisticated heuristics or excessive engineering.

\clearpage

\section*{Acknowledgment}
We would like to thank Manzil Zaheer, Angela Lin, Ashish Bora, and Thomas Crosley 
for their valuable comments and feedback on an early version of this paper.
This work was supported in part by Berkeley DeepDrive
and an equipment grant from Nvidia.

\bibliographystyle{splncs03}
\bibliography{../bib}

\begin{thebibliography}{10}
\providecommand{\url}[1]{\texttt{#1}}
\providecommand{\urlprefix}{URL }

\bibitem{uvg}
Ultra video group test sequences. \url{http://ultravideo.cs.tut.fi}, accessed:
  2018-03-11

\bibitem{vtl}
Video trace library. \url{http://trace.eas.asu.edu/index.html}, accessed:
  2018-03-11

\bibitem{webp}
Web{P}. \url{https://developers.google.com/speed/webp/}, accessed: 2018-03-11

\bibitem{agustsson2017soft}
Agustsson, E., Mentzer, F., Tschannen, M., Cavigelli, L., Timofte, R., Benini,
  L., Gool, L.V.: Soft-to-hard vector quantization for end-to-end learning
  compressible representations. In: NIPS (2017)

\bibitem{baig2017learning}
Baig, M.H., Koltun, V., Torresani, L.: Learning to inpaint for image
  compression. In: NIPS (2017)

\bibitem{balle2016end}
Ball{\'e}, J., Laparra, V., Simoncelli, E.P.: End-to-end optimized image
  compression. In: ICLR (2017)

\bibitem{carreira2017quo}
Carreira, J., Zisserman, A.: Quo vadis, action recognition? a new model and the
  kinetics dataset. In: CVPR (2017)

\bibitem{farneback2003two}
Farneb{\"a}ck, G.: Two-frame motion estimation based on polynomial expansion.
  In: SCIA (2003)

\bibitem{ioffe2015batch}
Ioffe, S., Szegedy, C.: Batch normalization: Accelerating deep network training
  by reducing internal covariate shift. In: ICML (2015)

\bibitem{jia2016dynamic}
Jia, X., De~Brabandere, B., Tuytelaars, T., Gool, L.V.: Dynamic filter
  networks. In: NIPS (2016)

\bibitem{jiang2017super}
Jiang, H., Sun, D., Jampani, V., Yang, M.H., Learned-Miller, E., Kautz, J.:
  Super slomo: High quality estimation of multiple intermediate frames for
  video interpolation. CVPR  (2018)

\bibitem{johnston2017improved}
Johnston, N., Vincent, D., Minnen, D., Covell, M., Singh, S., Chinen, T.,
  Hwang, S.J., Shor, J., Toderici, G.: Improved lossy image compression with
  priming and spatially adaptive bit rates for recurrent networks. arXiv
  preprint arXiv:1703.10114  (2017)

\bibitem{kingma2014adam}
Kingma, D.P., Ba, J.: Adam: A method for stochastic optimization. arXiv
  preprint arXiv:1412.6980  (2014)

\bibitem{le1991mpeg}
Le~Gall, D.: {MPEG}: A video compression standard for multimedia applications.
  Communications of the ACM  (1991)

\bibitem{lyty-vfsdv-17}
Liu, Z., Yeh, R., Tang, X., Liu, Y., Agarwala, A.: Video frame synthesis using
  deep voxel flow. In: ICCV (2017)

\bibitem{mcl-dmvpb-16}
Mathieu, M., Couprie, C., LeCun, Y.: Deep multi-scale video prediction beyond
  mean square error. In: ICLR (2016)

\bibitem{mentzer2018conditional}
Mentzer, F., Agustsson, E., Tschannen, M., Timofte, R., Van~Gool, L.:
  Conditional probability models for deep image compression. arXiv preprint
  arXiv:1801.04260  (2018)

\bibitem{cisco}
Networking~Index, C.V.: Forecast and methodology, 2016-2021. CISCO White paper
  (2016)

\bibitem{niklaus2017sepvideo}
Niklaus, S., Mai, L., Liu, F.: Video frame interpolation via adaptive separable
  convolution. In: ICCV (2017)

\bibitem{oord2016pixel}
Oord, A.v.d., Kalchbrenner, N., Kavukcuoglu, K.: Pixel recurrent neural
  networks. In: ICML (2016)

\bibitem{richardson2002video}
Richardson, I.E.: Video codec design: developing image and video compression
  systems. John Wiley \& Sons (2002)

\bibitem{waveone2017}
Rippel, O., Bourdev, L.: Real-time adaptive image compression. In: ICML (2017)

\bibitem{ronneberger2015u}
Ronneberger, O., Fischer, P., Brox, T.: U-net: Convolutional networks for
  biomedical image segmentation. In: MICCAI (2015)

\bibitem{schwarz2007overview}
Schwarz, H., Marpe, D., Wiegand, T.: Overview of the scalable video coding
  extension of the {H}.264/{AVC} standard. TCSVT  (2007)

\bibitem{Theis2017a}
Theis, L., Shi, W., Cunningham, A., Husz{\'a}r, F.: Lossy image compression
  with compressive autoencoders. In: ICLR (2017)

\bibitem{toderici2017full}
Toderici, G., Vincent, D., Johnston, N., Jin~Hwang, S., Minnen, D., Shor, J.,
  Covell, M.: Full resolution image compression with recurrent neural networks.
  In: CVPR (2017)

\bibitem{tsai2017learning}
Tsai, Y.H., Liu, M.Y., Sun, D., Yang, M.H., Kautz, J.: Learning binary residual
  representations for domain-specific video streaming. In: AAAI (2018)

\bibitem{vpt-gvsd-16}
Vondrick, C., Pirsiavash, H., Torralba, A.: Generating videos with scene
  dynamics. In: NIPS (2016)

\bibitem{wang2003multiscale}
Wang, Z., Simoncelli, E.P., Bovik, A.C.: Multiscale structural similarity for
  image quality assessment. In: ACSSC (2003)

\bibitem{w-ssgac-92}
Williams, R.J.: Simple statistical gradient-following algorithms for
  connectionist reinforcement learning. In: Reinforcement Learning. Springer
  (1992)

\bibitem{witten1987arithmetic}
Witten, I.H., Neal, R.M., Cleary, J.G.: Arithmetic coding for data compression.
  Communications of the ACM  (1987)

\bibitem{coviar}
Wu, C.Y., Zaheer, M., Hu, H., Manmatha, R., Smola, A.J., Kr{\"a}henb{\"u}hl,
  P.: Compressed video action recognition. In: CVPR (2018)

\bibitem{xue2016visual}
Xue, T., Wu, J., Bouman, K., Freeman, B.: Visual dynamics: Probabilistic future
  frame synthesis via cross convolutional networks. In: NIPS (2016)

\end{thebibliography}

\newpage
\noindent {\LARGE \textbf{Appendix}}
\appendix

\section{Model Details}
\paragraph{Context feature fusion.}
In experiments, we found that fusing the U-net features at resolution
$\frac{W}{2}\times \frac{H}{2}$ yields good performance for all models.
Fusing features at more
resolutions improves the performance slightly, but requires more memory and computation.
In this paper,
we additionally use $\frac{W}{4}\times \frac{H}{4}$ features
for the encoder of $\Mcal_{1,2}$ and $\Mcal_{3,3}$,
and $\frac{W}{4}\times \frac{H}{4}$ and
$\frac{W}{8}\times \frac{H}{8}$ features for the decoder of $\Mcal_{3,3}$.
Models are selected based on the performance on validation set.

\paragraph{Probability estimation in AAC.}
Adaptive arithmetic coding (AAC) relies on
a good probability estimation for each bit, given previously decoded bits, 
to efficiently encode a binary representation.
To estimate the probability,
we follow Mentzer~\etal
and use a 3D-Pixel-CNN model~\cite{mentzer2018conditional,oord2016pixel}.
The model contains 11 layers of masked convolution.
Each layer has 128 channels, and is followed by batch normalization~\cite{ioffe2015batch} and relu.
We train the models using Adam~\cite{kingma2014adam} with 
a learning rate of 0.0001 for 30K iterations.

\section{Bitrate Optimization}
We present detailed results of bitrate optimization.
\figref{beam_curves} shows the explored performance 
at the first level of the hierarchy.
We pick good combinations from the envelope of the curves,
and they proceed to the next level.
\figref{beam_results} presents the final combinations.
We use these combinations for the experiments in our paper unless otherwise noted.

\begin{figure}[t]
\center
\ra{1.4}
\begin{subfigure}[b]{0.52\textwidth}
      \begin{tikzpicture}
      \begin{axis}[
          legend pos=south east,
          width=1.2\linewidth,
          height=7cm,
          xlabel={BPP},
          every axis plot/.append style={ultra thick},
          cycle multiindex* list={
            mycolorlist
                \nextlist
            mystylelist
          },
      ]
      \pgfplotsset{every tick label/.append style={font=\scriptsize}}
      \addplot[only marks, chameleon2, mark=diamond*, mark size=3pt]
          table {plotdata/envelope.data};
          \addlegendentry{Picked $(K_0, K_1)$}
      \addplot
          table {plotdata/envelope1.data};
          \addlegendentry{$K_0=1$}
      \addplot
          table {plotdata/envelope2.data};
          \addlegendentry{$K_0=2$}
      \addplot
          table {plotdata/envelope3.data};
          \addlegendentry{$K_0=3$}
      \addplot
          table {plotdata/envelope4.data};
          \addlegendentry{$K_0=4$}
      \addplot
          table {plotdata/envelope5.data};
          \addlegendentry{$K_0=5$}
      \addplot
          table {plotdata/envelope6.data};
          \addlegendentry{$K_0=6$}
      \addplot
          table {plotdata/envelope7.data};
          \addlegendentry{$K_0=7$}
      \addplot
          table {plotdata/envelope8.data};
          \addlegendentry{$K_0=8$}
      \addplot
          table {plotdata/envelope9.data};
          \addlegendentry{$K_0=9$}
      \addplot
          table {plotdata/envelope10.data};
          \addlegendentry{$K_0=10$}
      \end{axis}
      \end{tikzpicture}\vspace{-2mm}
      \caption{MS-SSIM. Beam search for $K_1$ given $K_0$.}\label{fig:beam_curves}
\end{subfigure}\hfill
\begin{subfigure}[b]{0.4\textwidth}
  \setlength{\tabcolsep}{0.8em}
  \begin{tabular}{@{}lllll@{}}
    $K_0$&$K_1$&$K_2$&$K_3$&Average\\
    &&&&bitrate\\
    \midrule
    5 & 3 & 2 & 1 & 0.109\\
    7 & 3 & 2 & 2 & 0.151\\
    10 & 4 & 2 & 2 & 0.188\\
    10 & 10 & 2 & 2 & 0.219\\
    10 & 10 & 6 & 2 & 0.260\\
    10 & 10 & 10 &  2 & 0.302\\\\\\
  \end{tabular}
  \caption{Final combinations.}\label{fig:beam_results}
\end{subfigure}
\caption{Bitrate optimization. $K_0$, $K_1$, $K_2$ and $K_3$ 
correspond to the number of iterations used in 
the I-frame model, $\Mcal_{6,6}$, $\Mcal_{3,3}$ and $\Mcal_{1,2}$ respectively.
(a)~Beam search at the first level of hierarchy.
Picked $(K_0, K_1)$-combinations proceed to the next level.
(b) Final combinations returned by our algorithm. }\label{fig:beam}
\end{figure}

\end{document}